\documentclass{article}

\usepackage{arxiv}

\usepackage[utf8]{inputenc} % allow utf-8 input
\usepackage[T1]{fontenc}    % use 8-bit T1 fonts
\usepackage{hyperref}       % hyperlinks
\usepackage{url}            % simple URL typesetting
\usepackage{booktabs}       % professional-quality tables
\usepackage{amsfonts}       % blackboard math symbols
\usepackage{nicefrac}       % compact symbols for 1/2, etc.
\usepackage{microtype}      % microtypography
\usepackage{lipsum}
\usepackage{graphicx}
\graphicspath{ {./images/} }
\usepackage{color}

\usepackage[most]{tcolorbox}
\usepackage{enumitem}

\title{Assessing LLMs in Art Contexts: Critique Generation and Theory of Mind Evaluation}

\author{
 Takaya Arita \\
  Graduate School of Informatics\\
  Nagoya University\\
  Furo-cho, Chikusa-ku, Nagoya 464-8601, Japan \\
  \texttt{arita@nagoya-u.jp} \\
   \And
 Wenxian Zheng \\
  Graduate School of Informatics\\
  Nagoya University\\
  Furo-cho, Chikusa-ku, Nagoya 464-8601, Japan \\
  \texttt{bunnshoutei@gmail.com} \\
   \And
 Reiji Suzuki \\
  Graduate School of Informatics\\
  Nagoya University\\
  Furo-cho, Chikusa-ku, Nagoya 464-8601, Japan \\
  \texttt{reiji@nagoya-u.jp} \\
  \And
Fuminori Akiba \\
  Graduate School of Informatics\\
  Nagoya University\\
  Furo-cho, Chikusa-ku, Nagoya 464-8601, Japan \\
  \texttt{akiba.fuminori.f7@f.mail.nagoya-u.ac.jp} \\
}

\begin{document}
\maketitle
\begin{abstract}
This study explored how large language models (LLMs) perform in two areas related to art: writing critiques of artworks and reasoning about mental states (Theory of Mind, or ToM) in art-related situations. For the critique generation part, we built a system that combines Noël Carroll's evaluative framework with a broad selection of art criticism theories. The model was prompted to first write a full-length critique and then shorter, more coherent versions using a step-by-step prompting process. These AI-generated critiques were then compared with those written by human experts in a Turing test-style evaluation. In many cases, human subjects had difficulty telling which was which, and the results suggest that LLMs can produce critiques that are not only plausible in style but also rich in interpretation, as long as they are carefully guided. In the second part, we introduced new simple ToM tasks based on situations involving interpretation, emotion, and moral tension, which can appear in the context of art. These go beyond standard false-belief tests and allow for more complex, socially embedded forms of reasoning. We tested 41 recent LLMs and found that their performance varied across tasks and models. In particular, tasks that involved affective or ambiguous situations tended to reveal clearer differences. Taken together, these results help clarify how LLMs respond to complex interpretative challenges, revealing both their cognitive limitations and potential. While our findings do not directly contradict the so-called Generative AI Paradox—the idea that LLMs can produce expert-like output without genuine understanding—they suggest that, depending on how LLMs are instructed, such as through carefully designed prompts, these models may begin to show behaviors that resemble understanding more closely than we might assume. 
\end{abstract}

% keywords can be removed

\section{Introduction}
\begin{figure} [t] % picture
    \centering
    \includegraphics[width=60mm]{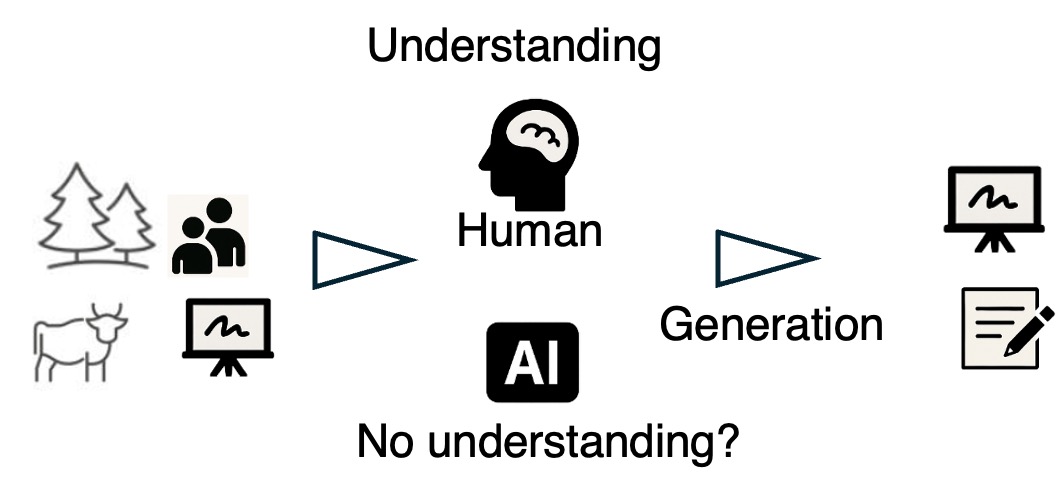}
  \caption{Generative AI paradox.}
  \label{fig:fig1}    
\end{figure}

The rapid development of artificial intelligence, particularly large language models (LLMs), has brought about striking advances in content generation. These models are now capable of producing texts, images, and music that are often indistinguishable from those created by humans and, in some cases, even surpass them in fluency or technical polish. Yet, this generative capability reveals a fundamental asymmetry. Human creativity is typically grounded in understanding, where expertise in generating content arises from deep conceptual knowledge. In contrast, LLMs appear to produce convincing outputs without such understanding. This disconnect between generation and understanding is referred to as the Generative AI Paradox \cite{west2023generative} (Figure \ref{fig:fig1}).

\begin{figure} [t] % picture
    \centering
    \includegraphics[width=120mm]{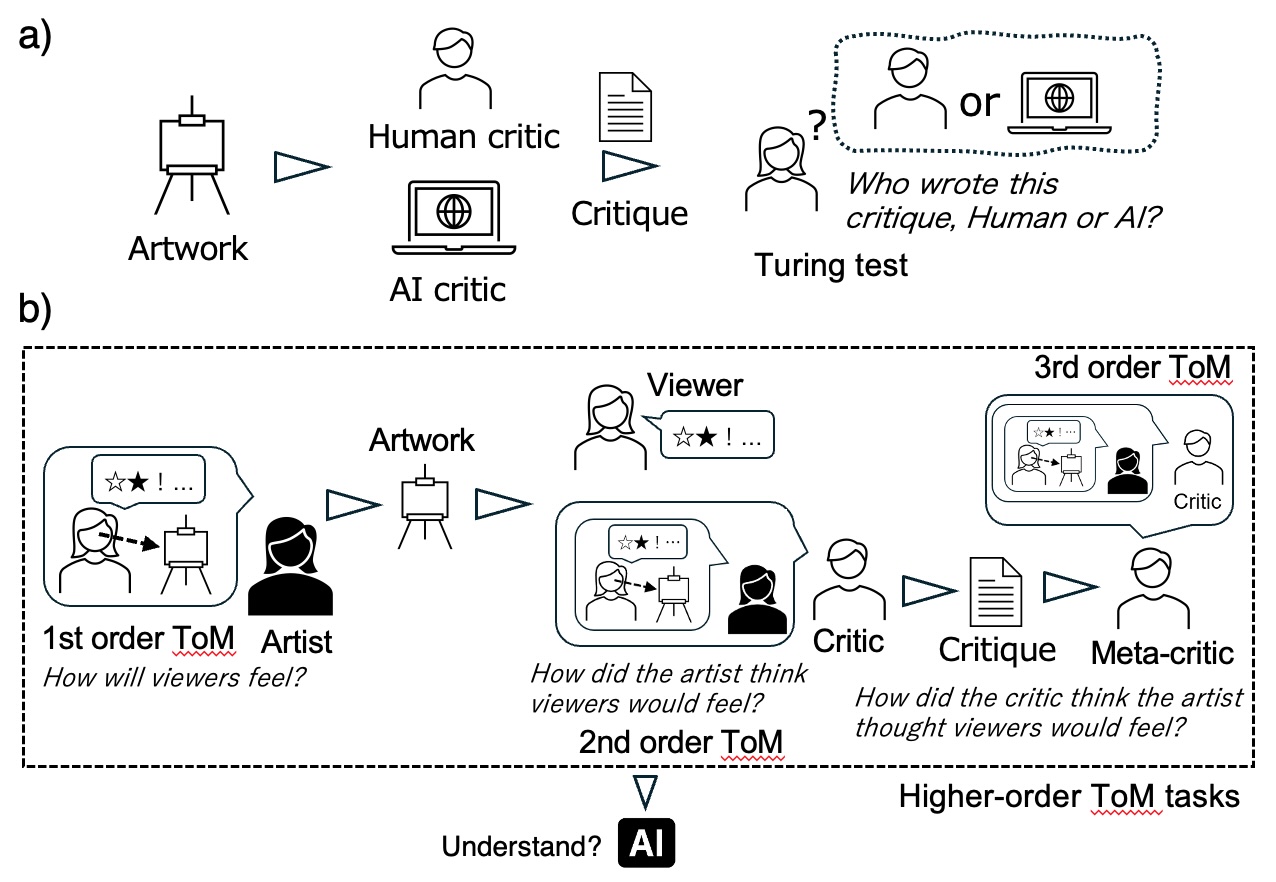}
  \caption{Two art contexts to assess LLMs: a) Critique generation and b) Theory of Mind evaluation.}
  \label{fig:fig2}    
\end{figure}

This study carefully examines this paradox in the following two art contexts to provide a foundation for further consideration (Figure \ref{fig:fig2}). First, we will examine whether it is possible for LLMs to generate in-depth art critiques from a perspective grounded in aesthetic theory. To evaluate this, we will conduct a Turing test specialized in art criticism. Second, we will examine whether LLMs have a higher-order version of the Theory of Mind (ToM) \cite{premack1978does} that may be involved when writing and reading a critique. To do this, we propose three ToM tasks and evaluate a large number of LLMs using these tasks.

Before discussing the two approaches, it is important to clarify our stance on the evaluation of artworks. First, the answer to the question of how to treat the value of artwork in criticism can generally be divided into two positions: one that does not go into value judgment but only interpretation and one that holds that value judgment is important. According to Gombrich, comprehending a work of art necessitates an understanding of its historical, cultural, and social context \cite{Gombrich1960-GOMAAI}. In addition, Danto advanced the concept of the art world to assert that the significance of an artwork is determined by its interpretation within the context of the \textit{artworld} \cite{danto1964artworld}. According to Danto, criticism should prioritize interpreting the meaning and concept of the work rather than evaluating its value. This perspective seems to have dominated the second half of the 20th century. 

In contrast, Carroll contends that the critic's primary task is value judgment: to discover, articulate, and justify the value of the work, which serves as a guide for the viewer to appreciate artworks more deeply \cite{carroll2009criticism}. Furthermore, Carroll assesses the significance of \textit{success value}, which prioritizes the extent to which artworks have achieved their intended goals and conceptual underpinnings, as opposed to \textit{reception value}, which is contingent on the viewer's emotional responses, interpretations, and experiences. He further contends that it is only through the adoption of success value as the foundation for evaluation that we can adequately assess the originality and historical significance of an artwork.

In this study, the two approaches previously mentioned will be designed and conducted in accordance with Carroll's position that evaluation should accompany criticism, and that this evaluation should emphasize success value. We believe that it leads to a more profound investigation into the capabilities of LLMs in the artistic domain and contributes to the development of more sophisticated LLMs, as well as to our understanding of human creativity.

It might be important to note that the value of an artwork is not solely determined by the communication between the artist and the viewer through the artwork or by critics alone. The determination of a work's value is also influenced by institutions such as museums, the art market, and curators, among others. Moreover, it is subject to modification in accordance with shifts in the cultural and social context. In light of these considerations, we have adopted a specific stance in our approaches to evaluate LLMs in the art contexts.

\section{Generating artwork critiques using LLM}
\subsection{Background}
In recent years, the rapid progress of artificial intelligence has led to significant advances in research on AI-based image captioning. Nevertheless, these are basically aimed at generating descriptions based on formal analysis, often within a few sentences, and have not yet reached the point of generating aesthetically grounded art criticism as written by human art experts. Looking back at previous research on semantic art understanding, in a seminal paper, Garcia and Vogiatzis presented SemArt, the first dataset of fine art images paired with corresponding artistic commentaries including background information and the historical context of the work \cite{garcia2018read}. They conducted several experiments using this dataset, although the aim was not so much to generate art criticism as to pave the way for further exploration in the field. More recently, Cetinic extended this line of research by generating iconographic captions using a transformer-based model trained on a large-scale dataset annotated with Iconclass codes, targeting deeper semantic associations \cite{cetinic2021towards}. Evaluation includes traditional metrics and CLIPScore \cite{hessel2021clipscore}, demonstrating that the model generates captions more relevant to artworks than those trained on natural image datasets.

With the rise of LLMs in recent years, more sophisticated systems have demonstrated the ability to generate detailed, visually grounded formal analyses of artworks. GalleryGPT \cite{bin2024gallerygpt} employs a fine-tuned LLM to focus on elements such as composition, light, and color, tying critiques directly to visual content and avoiding reliance on external metadata or pre-existing textual knowledge. They also identified the problem of "LLM-biased visual hallucination," where models tend to recognize famous artworks and generate critiques based on memorized data rather than genuine visual analysis. Their work inspired us to implement safeguards to mitigate the risk of AI-generating prelearned outputs. ArtGPT-4 incorporates specialized adapter layers into a significantly smaller LLM compared to state-of-the-art models like GPT-4, enabling it to analyze fine details in artistic images and enhance emotional understanding \cite{yuan2023artgpt}. It is stated that ArtGPT-4 is capable of discerning and articulating the emotions elicited by an image like a human being, such as feelings of positivity and inspiration. The study by Castellano and others \cite{castellano2023exploring} automatically generates visual descriptions of artworks using ChatGPT, filters and weights them using CLIPScore, and then fine-tunes a pre-trained image-to-text model with this data to produce captions that describe the artworks' visual content. They argue that conventional AI-generated captions for artworks lack an understanding of artistic intent, historical background, and understanding of complex visual elements. However, the training data is biased toward visual descriptions, the model is not necessarily designed to incorporate artistic interpretation, and CLIPScore is not well-suited for evaluating symbolic meaning.

\subsection{Theoretical foundation}
\subsubsection{Noël Carroll's criticism framework}
We adopt Noël Carroll's theory of criticism \cite{carroll2009criticism} as the theoretical foundation for designing a method for creating art critiques by LLM. Carroll's framework integrates seven components—\textbf{description}, \textbf{classification}, \textbf{contextualization}, \textbf{elucidation}, \textbf{interpretation}, \textbf{analysis}, and \textbf{evaluation}—all working together to create a coherent and well-structured critique. While the first six elements serve distinct roles in understanding and explaining an artwork, they ultimately support the central task of evaluation. The balance and emphasis on these supporting elements naturally vary depending on the genre and nature of the artwork in question, ensuring that each critique remains tailored to the specific work. Below are the specific procedures for the six supporting components:

\begin{enumerate}[label=\arabic*)]
\item \textbf{Description}: Identifying and detailing the observable features of the artwork, such as its color, composition, technique, medium, and subject matter. This step provides the foundation for further analysis and interpretation.
\item \textbf{Classification}: Placing the artwork within an appropriate category, such as genre, style, movement, or medium, to establish expectations and provide a framework for evaluation.
\item \textbf{Contextualization}: Situating the artwork within its historical, cultural, and social environment, as well as considering the personal background of the artist, to uncover deeper meanings and significance.
\item \textbf{Elucidation}: Clarifying the symbolic, semantic, and iconographic meanings within the artwork, often by identifying icons, motifs, and culturally significant symbols.
\item \textbf{Interpretation}: Examining the artwork on a broader scale, including its overarching themes, narrative actions, and underlying messages to reveal its significance. 
\item \textbf{Analysis}: Explaining how various elements within the artwork contribute to its coherence, unity, or intentional disunity, often by exploring their emotional or aesthetic impact.
\end{enumerate}

Appendix A shows the file given to the LLM as external knowledge of Noël Carroll's framework of criticism.

\subsubsection{Various theories of art criticism}
Art criticism draws on a wide variety of theoretical perspectives. We employ 15 distinct theories of criticism summarized in Ogura's comprehensive list \cite{ogura2023for} to form the basis of the AI-generated critiques: \textbf{Structuralist Criticism}, \textbf{Narrative Criticism}, \textbf{Reception Theory Criticism}, \textbf{Deconstructionist Criticism}, \textbf{Psychoanalytic Criticism}, \textbf{Thematic Criticism}, \textbf{Feminist Criticism}, \textbf{Gender Criticism}, \textbf{Genetic Criticism}, \textbf{Marxist Criticism}, \textbf{Cultural Materialist Criticism}/\textbf{New Historicist Criticism}, \textbf{Socio-criticism}, \textbf{Cultural Studies}, \textbf{Systems Theory Criticism}, and \textbf{Postcolonial Criticism}/\textbf{Transnationalism}.

Each theory provides a unique lens through which to perform Noël Carroll's steps of criticism, guiding the critic to focus on particular aspects of a work's meaning or construction. As an example of Feminist Criticism, consider Emily Osborne's \textit{Nameless and Friendless} \cite{pollitt2015osborn}, which was later used in the Turing test. The painting captures a young, vulnerable woman attempting to sell her art in a cold, male-dominated society. The surrounding figures' indifferent or dismissive expressions highlight the isolation and systemic gender inequality faced by women in Victorian England. This is consistent with Carroll's focus on contextualization, as the work becomes a window into the societal norms of its time. This example illustrates how each of those theories provides distinct yet complementary insights that enrich each stage of Carroll's framework.

Appendix B shows the file given to the LLM as external knowledge of the 15 distinct theories of criticism.

\subsection{Implementation}
\subsubsection{External knowledge}
\begin{figure} [htbp] % picture
    \centering
    \includegraphics[width=80mm]{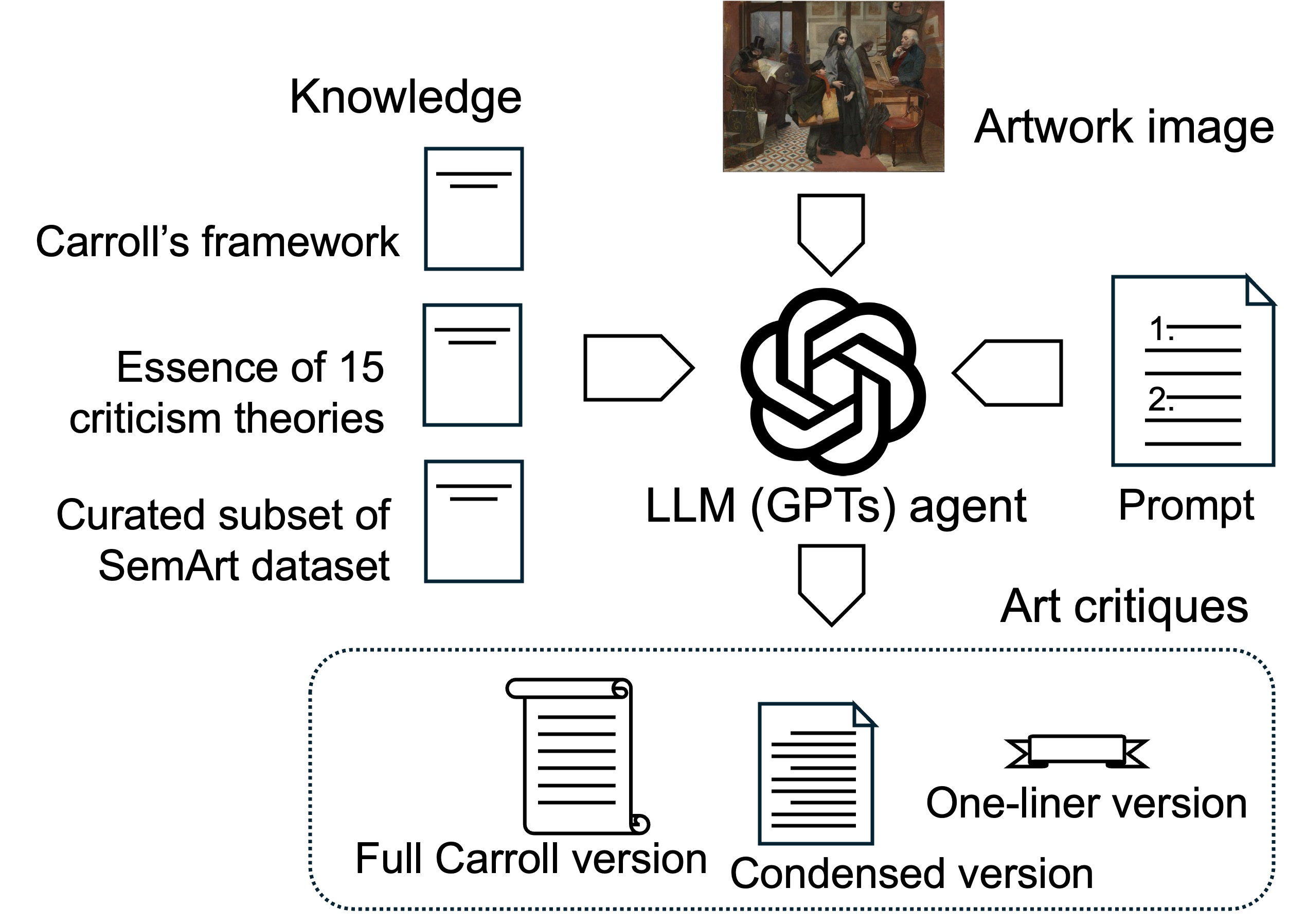}
  \caption{Configuration of Composer.}
  \label{fig:fig3}    
\end{figure}

We have developed \textbf{Composer}, a system that utilizes a custom GPT built with OpenAI's GPTs framework. This framework allows developers to configure GPT-4o\footnote{According to the official GPT-4o page (https://platform.openai.com/docs/models/gpt-4o), the version used is presumed to be gpt-4o-2024-08-06, based on the usage date.}-based models with specific instructions, external knowledge, and tools, enabling them to perform specialized tasks such as structured art critique generation. The configuration of Composer is shown in Figure  \ref{fig:fig3}. To equip Composer with the foundational knowledge required for Carroll's framework of criticism, we uploaded a self-authored summary of Noël Carroll's book On Criticism \cite{carroll2009criticism}. This summary delineates the key elements of each step in Carroll's methodology, providing detailed guidance and concrete examples. By focusing on these essential components, the summary ensures the model can generate critiques that align with Carroll's structured and academic approach. Similarly, we uploaded a summary of the above 15 theories of criticism to provide the model with diverse theoretical perspectives. This summary contains about three sentences that describe the essence of each critical theory. Together, these resources enable the model to generate nuanced and well-rounded critiques grounded in both structured methodology and diverse theoretical viewpoints. Furthermore, we semi-automatically selected 532 caption-style descriptions from the SemArt dataset, based on content and volume, to help the model learn the phrases and linguistic styles commonly found in art-related captions, albeit not full critiques\footnote{Available at https://nuss.nagoya-u.ac.jp/s/mL2k95JFxNRKnb2}. While the specific impact of this textual input has not been systematically evaluated, it was included as a supplementary resource to support stylistic fluency. Together, these resources enable LLMs to learn to generate nuanced and well-rounded critiques from methodological, theoretical, and practical viewpoints.

\subsubsection{Prompts and output refinement}
Through an iterative improvement process, we developed a prompt (Appendix C) that guides Composer in generating critiques with a structure that conforms to Carroll's framework, based on insights from various theories of art criticism. It instructs Composer to begin with a comprehensive, full-length critique. This version consists of all seven components according to Carroll's method. Each component is rigorously defined and associated with relevant theories, ensuring that every potential insight is articulated. This approach produces analytically rich content. However, even if each analysis and interpretation seem reasonable, there is a possibility that the overall critique is not cohesive or that it is over-interpreted in parts.

The second step is to produce a condensed three-paragraph version summarizing the full-length version. The purpose of this condensed version is to reorganize and rephrase the content to improve readability, coherence, and natural expression. The critiques of this version will be used for evaluation, as described in the next section. Finally, Composer generates a one-liner critique to distill the essence of the artwork into a short, playful commentary. These one-liners aim to capture the spirit of the artwork in an engaging and memorable way, drawing attention to key features while remaining concise and impactful. While primarily intended for stylistic exploration, one such one-liner was also used in the Turing test as a supplementary example.

This refinement workflow mirrors the \textit{chain-of-thought} prompting technique \cite{wei2022chain}, where the first version serves as a detailed groundwork for subsequent editions. By extracting and condensing key information, Composer transitions from generating raw, detailed observations to producing polished, human-like critiques.

Appendix D shows examples of these three versions of the critiques generated.

\section{Turing test on the generated critiques}
\subsection{Background}
In his seminal paper \cite{turing1950computing}, Alan Turing proposed what is now called the Turing test, in which a human judge engages in text-based conversations with both a human and an AI without knowing which is which. If the machine can imitate the human closely enough that the judge cannot reliably tell them apart, it is said to have passed the test.

In 2014, during an event at the Royal Society in London, a chatbot named Eugene Goostman was reported to have passed a version of the test by convincing 33\% of human judges that it was human \cite{warwick2016can}. Its success was attributed to several clever strategies: adopting the persona of a 13-year-old Ukrainian boy, frequently shifting topics or asking questions back to the judge, incorporating humor, and inserting typos at random. These techniques helped mask its limitations and gave plausible reasons for linguistic oddities or knowledge gaps.

While the validity of this result remains debated—since the setup somewhat diverged from Turing's original protocol—it nevertheless marked a cultural milestone in public perception of AI. In the years since the rapid development of LLMs has led to numerous Turing-style experiments \cite{jones2024does} \cite{jones2024people} \cite{rathi2024gpt}. Some of these adopt simplified formats where human subjects are presented with a single response and asked to judge whether it is written by a human or an AI rather than choosing between two entities. Most recently, a three-party Turing test by Jones and Bergen demonstrated for the first time that a modern LLM (GPT-4.5) could not only pass the original formulation of the test but was even more likely to be judged as human than the actual human participant when prompted with a suitable persona \cite{jones2025large}.

Considering the above, it seems that the historical role of the Turing test, which measures the degree of imitation of conversation as a measure of intelligence, is coming to an end and that a new development period is approaching. In this study, we use the Turing test to focus on the differences between humans and AI in terms of understanding, interpretation, and the evaluation of creativity, which is often regarded as a core aspect of what distinguishes human intelligence, and the basis for this evaluation, rather than superficial linguistic style.

\subsection{Test design and procedure}
\subsubsection{System design}
\begin{figure} [htbp] % picture
    \centering
    \includegraphics[width=130mm]{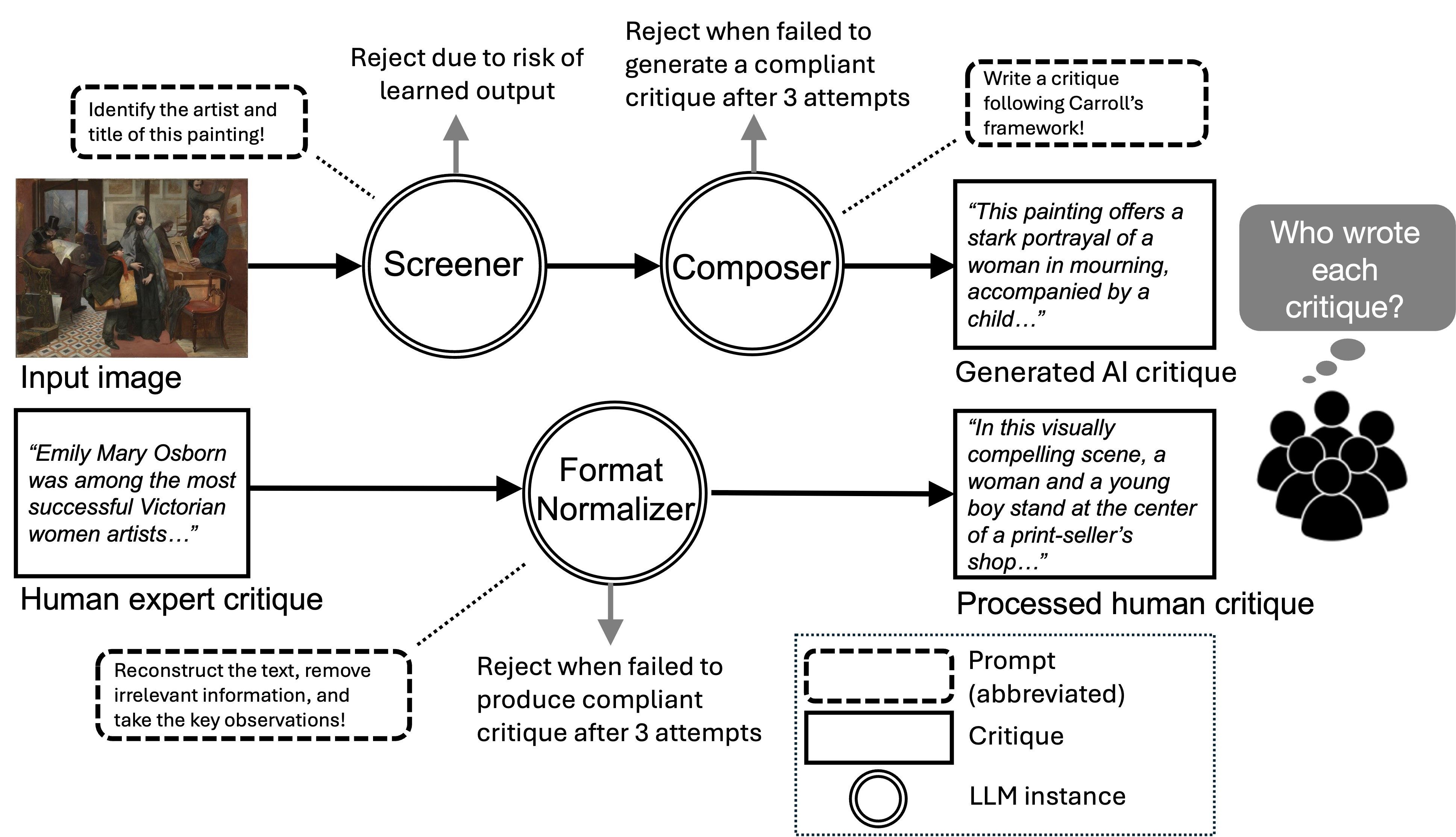}
  \caption{Workflow for Turing test data preparation.}
  \label{fig:fig4}    
\end{figure}

This subsection outlines the methodology employed to generate, refine, and evaluate art critiques (which serve as stimuli in the Turing test) using three distinct instances of LLMs (Figure \ref{fig:fig4}). For clarity, we designate specific names to these instances based on their roles: \textbf{Screener}, which identifies and removes artworks that are too well-known to ensure a fair comparison; \textbf{Composer} (described in the previous section), which utilizes external knowledge and a carefully crafted prompt to generate original critique; and \textbf{Format Normalizer}, which processes critiques written by human experts to align their format with AI-generated outputs for comparability.

Together, these instances produce sets of test data, each consisting of an input image, an original AI critique, and a processed human critique. We also used the same customized GPT-4o model developed with OpenAI's GPTs framework, as described in Section 2.3.1. We believe that the observations and conclusions drawn here can be reasonably extended to other LLMs. The prompts for Screener, Composer, and Format Normalizer are shown in Appendix E.

To ensure that the Turing test is fair, we individually check each generated data pair against the following rules:

%\begin{enumerate}[label=\textbf{R\arabic*}\,:]
\begin{enumerate}[label=\textbf{R\arabic*.}]
  \item Generated critiques must not rely on or reproduce pre-existing text written by humans, which ensures their originality. This setting is based on two reasons: to investigate how deeply LLM can interpret an artwork without directly using human written critiques and to generate sophisticated critiques even for entirely new artworks.
  \item Generated critiques must be free of obvious factual errors. Every claim within the critique is checked to maintain credibility and quality.
  \item Generated critiques must not include facts that cannot be reasonably inferred from the visual features of the artwork. For example, details such as the artist's name, the exact year of creation, or unobservable historical context must be excluded.
  \item The critiques must be consistent in length and style. This helps ensure that variations in presentation do not bias the subjects' judgments.
  \item Both critiques must include theoretical considerations related to the features of the artwork. These considerations should offer depth and analytical insight.
\end{enumerate}

\subsubsection{Human critique processing}
To prepare human critique counterparts for the Turing test, we sourced comprehensive critiques from Smarthistory\footnote{https://smarthistory.org}, a nonprofit organization dedicated to art education. One of Smarthistory's primary missions is to provide educational materials for students in art classes. Its extensive database of peer-reviewed critiques, written by art experts, ensures a high standard of quality and the credibility of human expertise. Nevertheless, Smarthistory's informative contents needed to be adapted to the structure and goals of our study. To achieve this, we employed Format Normalizer, which performs three tasks: First, it removes or masks any information that cannot be directly inferred from the visual features of the artwork (\textbf{R3}). Second, it ensures that the critique remains faithful to Smarthistory's original content by preserving the reviewer's wording and retaining key observations, analyses, and thoughts (\textbf{R5}). Third, it reorganizes the original text into a standardized three-paragraph format to improve readability (\textbf{R4}). By following these instructions, Format Normalizer ensures that critiques are focused on observable features such as composition, color, and symbolism while retaining essential analytical depth. 

\subsubsection{Dataset construction}
The following steps outline the procedure used to prepare the Turing test, ensuring compliance with the rules and criteria established.

\begin{enumerate}[label=\arabic*)]

\item Selection of artworks: We selected artworks from Smarthistory that come with comprehensive human-authored critiques. The selected artworks are mostly European classics, including still life and portraits.
\item Filtering out AI-known artworks (\textbf{R1}): The artwork image was uploaded to Screener. If Screener provided incorrect answers or failed to confirm either the title or artist, the artwork was selected for further processing. Otherwise, it was excluded.
\item Generating AI critiques (\textbf{R4} and \textbf{R5}): Composer generates three critiques for each input image: a full-length comprehensive critique, a condensed critique, and a one-liner, as described in 2.3.
\item Ensuring accuracy in AI critiques (\textbf{R2}): Each generated critique was carefully reviewed for obvious factual errors. If inaccuracies were found, a new session was initiated, and the critique was regenerated. If Composer failed to generate an accurate critique after three attempts, the artwork was discarded.
\item Processing human critiques (\textbf{R3}): The human-authored critique was processed by Format Normalizer. The processed critiques were carefully reviewed to ensure compliance with \textbf{R3}. If the output contained prohibited content, a new session was initiated, and the text was regenerated. If Format Normalizer failed to produce a compliant critique after three attempts, the artwork was discarded.
\item Finalizing Turing test data: For each successfully processed artwork, a dataset was created comprising the following elements: the artwork image, basic information about the artwork, the generated critique, and the processed human critique. These datasets were presented to the subjects for identification.

\end{enumerate}

\subsubsection{Test procedure}
The following five artworks were the subject of critiques for the Turing test, with a human and LLM-generated critique for each (Appendix F):

\begin{enumerate}[label=\arabic*)]

\item \textit{Still Life with Ebony Chest} (1652) by Antonio de Pereda
\item \textit{A View of Paris: The Life of Pleasure} (1944) by Jean Du-buffet
\item \textit{The Jewish Cemetery} (1655) by Jacob van Ruisdael
\item \textit{The Family of Carlos IV} (1801) by Francisco Goya
\item \textit{Nameless and Friendless} (1857) by Emily Mary Osborn

\end{enumerate}

All but 4) used condensed three-paragraph generated critiques, which were compared to the processed human critiques. For 4), a one-liner critique was used, compared to the renowned satirical critique often attributed to Théophile Gautier: "\textit{A picture of the corner grocer who has just won the lottery}". Since this criticism appears to be completely off the mark, it would be difficult to judge that it was written by a human being unless you already knew about the criticism (which none of the human subjects were familiar with). Also, because it is exceptional in terms of being a single-sentence critique, we have decided to include statistical values both with and without the test results for this work, due to its exceptional nature as a one-liner critique.

In addition to judging which of the two critiques was written by a human, the subjects also answered a questionnaire: for each work, they were asked to rate their confidence in their judgment on a five-point scale, to give their reasons for their judgment, and to select which of the two critiques they preferred, regardless of which one they had chosen. They were also asked to report their affinity for art on a five-point scale (the extent of their knowledge of art and how often they visited exhibitions) and the themes of art criticism that interested them.

Our analysis is focused on three main questions, which are discussed in Subsections \textbf{3.3.1}, \textbf{3.3.2}, and \textbf{3.3.3}, respectively.

\begin{enumerate}[label=\arabic*)]
%\begin{enumerate}[label=\textbf{\Alph*.}]
  \item Accuracy of Identification: How accurately could subjects distinguish generated critiques from human-authored ones? 
  \item Factors Related to Accuracy: What factors influenced identification accuracy? This includes: Confidence, the correlation between subjects' reported confidence levels and their identification accuracy; Art Knowledge, whether subjects with greater art-related experience were more likely to make accurate judgments; and Preference, whether the critique a participant preferred was also the one they identified as human-authored, and how this relationship differed between correct and incorrect responses.
  \item Reasons for Identification: What specific features or justifications did subjects cite as the basis for their decisions? This question provided qualitative insights into perceived differences that shaped their judgments.
\end{enumerate}

\subsection{Results}
The Turing test took place in October 2024 as part of Science Agora\footnote{https://www.jst.go.jp/sis/scienceagora/}, a national forum for science communication and public engagement. The people who visited our booth and took the test were individuals with an apparent interest in the arts (high school age and above), and 60 valid responses were obtained. In the following, Q1 to Q5 correspond to the above artworks 1) to 5).

\subsubsection{Accuracy of identification}
\begin{figure} [htbp] % picture
    \centering
    \includegraphics[width=90mm]{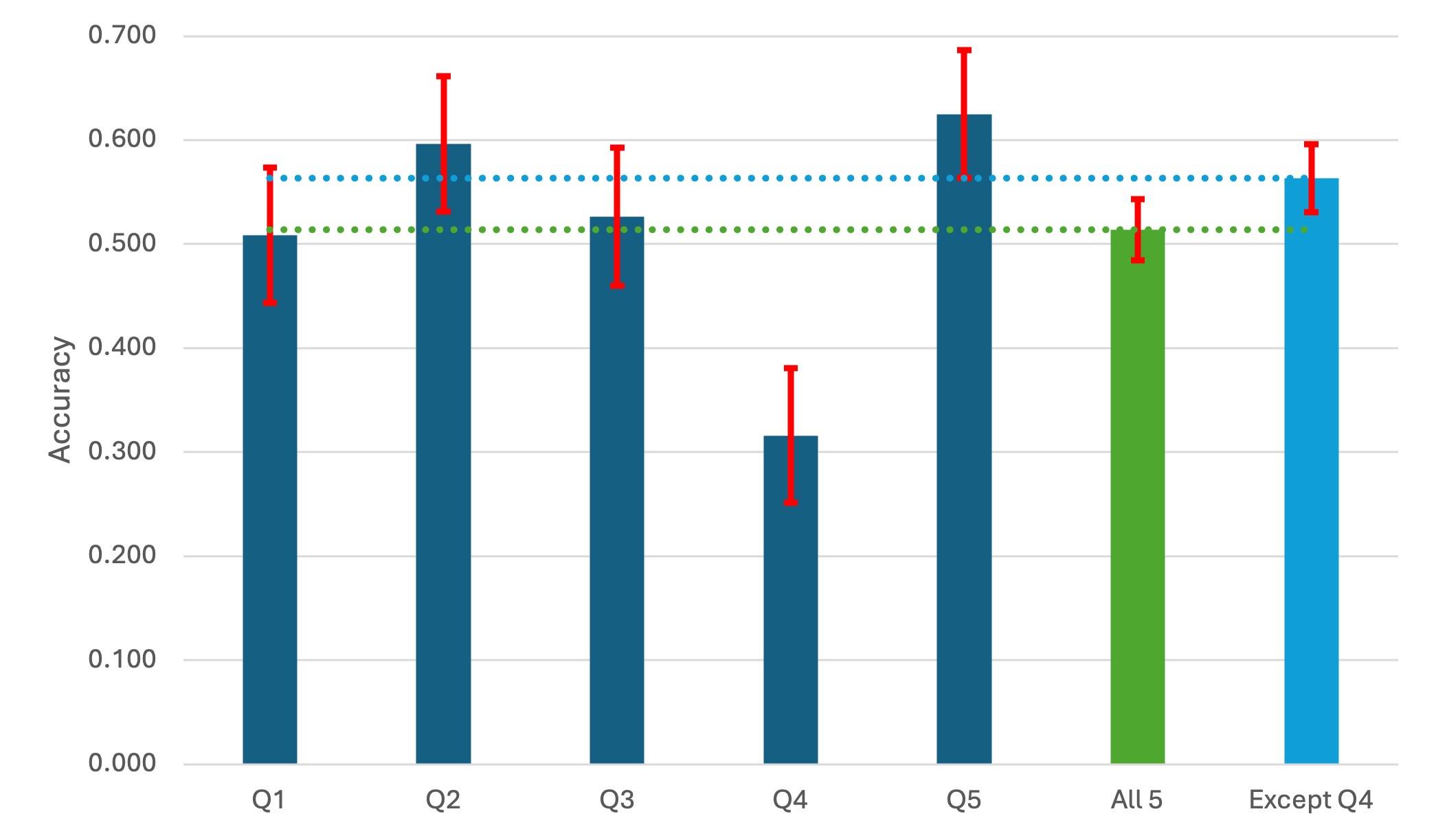}
  \caption{Accuracy in identifying human critiques.}
  \label{fig:fig5}    
\end{figure}

Figure \ref{fig:fig5} shows the overall accuracy results. The error bars represent ±1 standard error of the mean (SE), applied consistently in all figures where error bars are shown. This result shows that the human subjects performed only slightly better than random guessing, with an average accuracy of 51.4\%, which is only marginally above random chance. When excluding the results of Q4—an outlier in both format and outcome—the accuracy increases to 56.3\%. According to Turing's original proposal, an AI is considered to have passed the Turing test if it consistently convinces at least 30\% of subjects that its responses are human-authored. In this sense, both results comfortably meet Turing's original benchmark of 30\% for passing the test. In particular, many subjects struggled with the subtle, nuanced critique written by Théophile Gautier in Q4. They often mistook the apparent mismatch between the critique and the painting as a failure of visual recognition of LLM, opting instead for the AI-generated critique, which seemed more straightforward. This contributed to the significantly lower accuracy observed for Q4. The satire may have been too human for its own good. Ironically, this led not a few participants to suspect it was written by an AI.

The binomial test yielded the following p-values for each Turing test question: Q1 (N = 59, p = 1.000), Q2 (N = 57, p = 0.185), Q3 (N = 57, p = 0.791), Q4 (N = 57, p = 0.008), and Q5 (N = 56, p = 0.081). Only Q4 showed a significant deviation from chance, with subjects misidentifying the AI-generated critique as human-authored. Overall, in most cases, subjects were unable to reliably distinguish between generated critiques and human-authored critiques, with their judgments appearing to approximate random guesswork. 

\subsubsection{Factors related to accuracy}

\begin{figure} [htbp] % picture
    \centering
    \includegraphics[width=90mm]{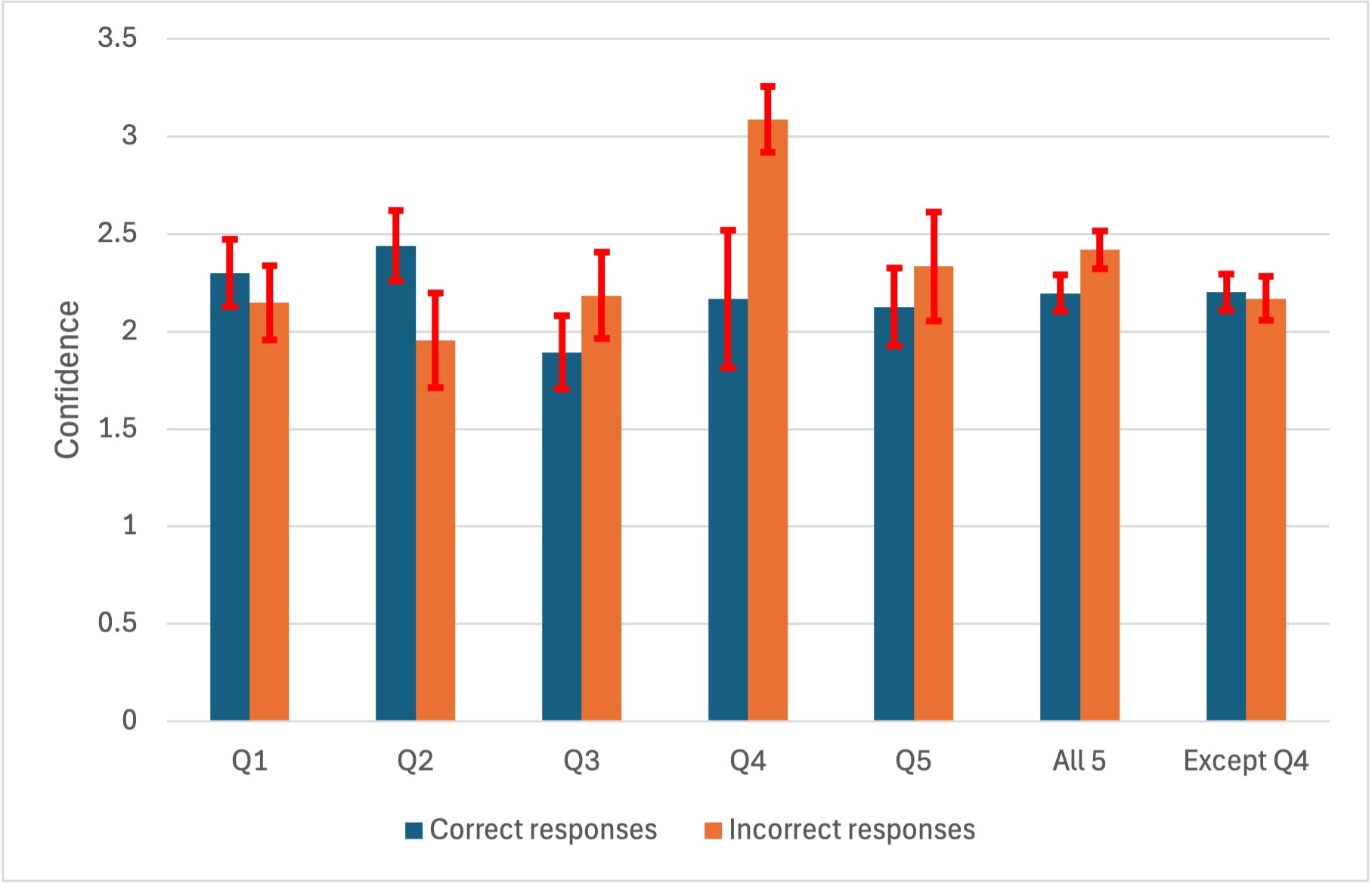}
  \caption{Confidence levels by identification accuracy.}
  \label{fig:fig6}    
\end{figure}

Figure~\ref{fig:fig6} shows the confidence levels for correct and incorrect responses across all five Turing test questions. An analysis of these confidence levels revealed no significant differences between correct and incorrect responses for any of the questions, except Q4. The average confidence for correct responses was 2.20, while the average for incorrect responses was slightly higher at 2.42. When excluding Q4, the difference narrowed further to 2.20 versus 2.17. Obviously, Q4 stands out as an exception, with human subjects who provided incorrect responses exhibiting higher confidence (3.09, SE = 0.169) compared to those who answered correctly (2.17, SE = 0.355). This indicates that subjects mistakenly found the AI-generated critique in Q4 more convincing, potentially due to the apparent quality of the generated critique.

\begin{figure} [htbp] % picture
    \centering
    \includegraphics[width=90mm]{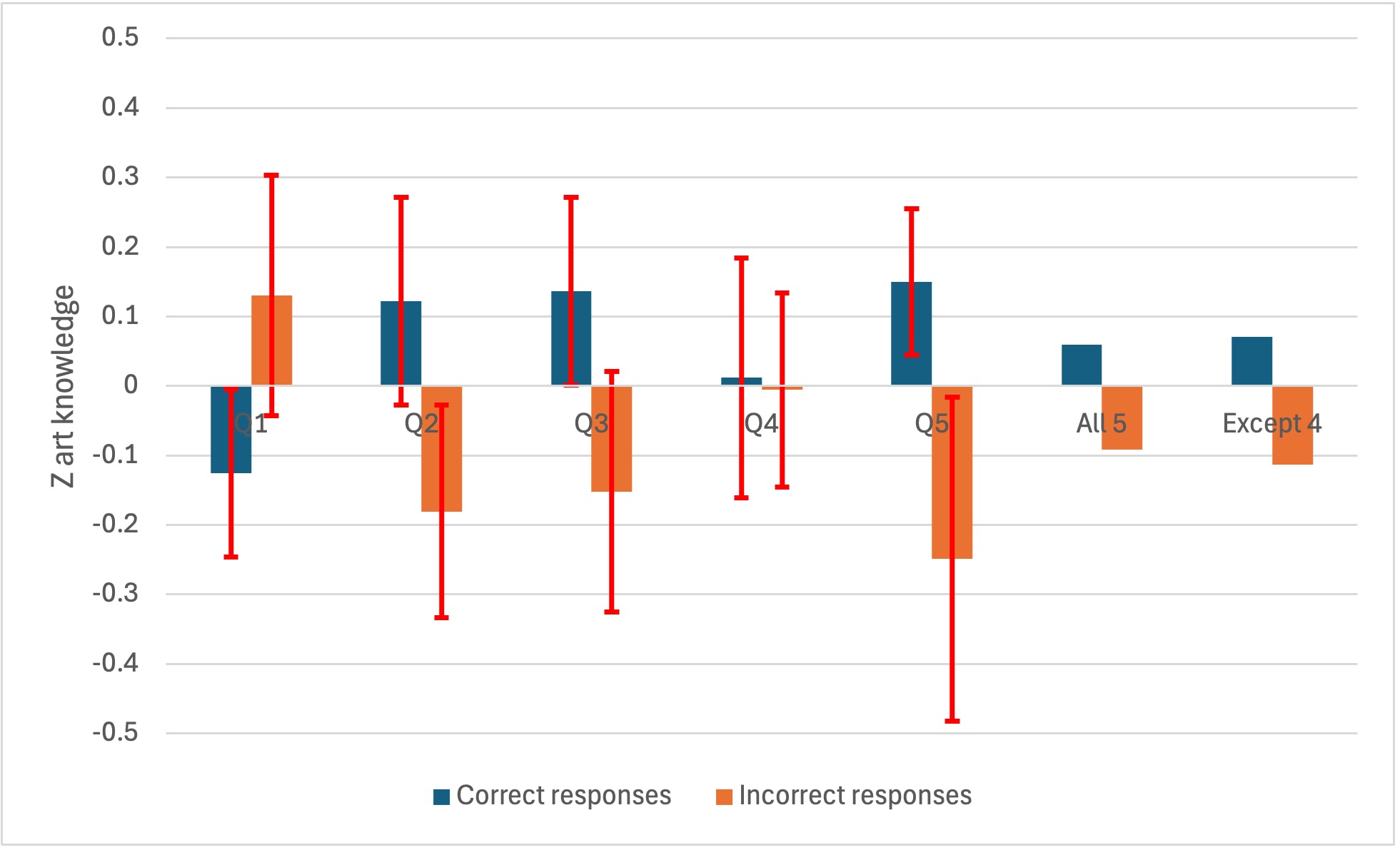}
  \caption{Standardized art knowledge scores by identification accuracy.}
  \label{fig:fig7}    
\end{figure}

\begin{figure} [htbp] % picture
    \centering
    \includegraphics[width=90mm]{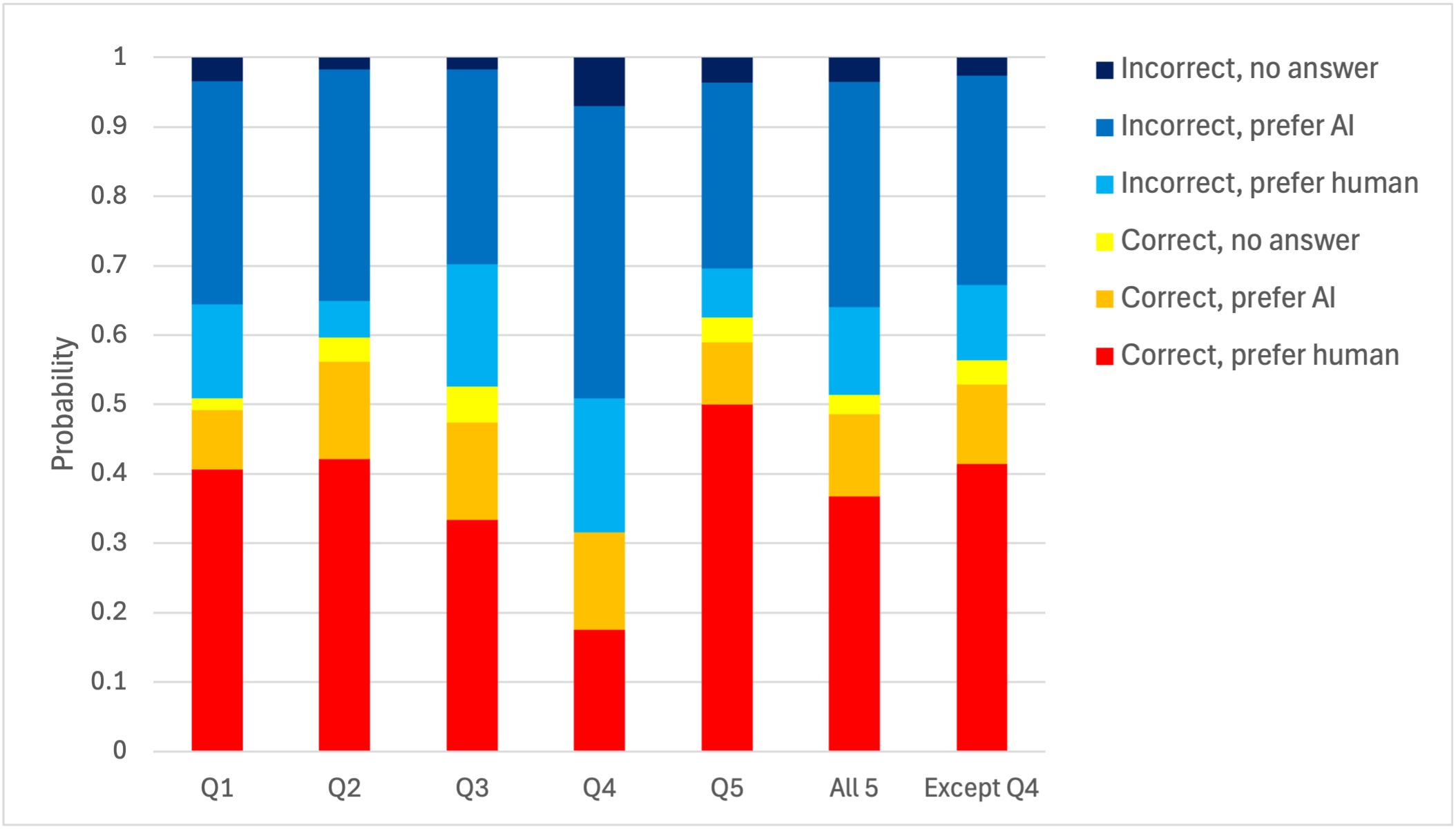}
  \caption{Preference patterns by identification accuracy.}
  \label{fig:fig8}    
\end{figure}

We then analyzed the self-reported art knowledge scores, which ranged from 1 (rarely visit art galleries, limited knowledge of art) to 5 (frequent gallery visits, extensive understanding of art). These scores were standardized using z-score normalization (subtracting the population mean and dividing by the standard deviation). The results are summarized in Figure \ref{fig:fig7}, which presents a breakdown of how subjects' critique preferences align (or not) with their ability to correctly identify the human-authored critique. The stacked bar chart shows proportions of combinations such as "correct and preferred human critique" (in red), as well as other patterns of correctness and preference. We observed that, for most questions, the average art knowledge score was higher among subjects who gave correct responses compared to those who responded incorrectly, with the exception of Q1, for which the reason remains unclear. However, the differences between groups were not statistically significant, as the confidence intervals for correct and incorrect responses substantially overlapped and did not indicate consistent patterns. This may be partly due to the multifaceted nature of aesthetic judgment, which cannot be fully accounted for by knowledge level alone. Subjective factors such as interpretation style, emotional resonance, and personal preference likely play a significant role.

Another important aspect of our analysis was the examination of subjects' preferences for the critiques, regardless of their identification accuracy. After identifying which critique they believed was AI-generated, subjects were asked to select their preferred critique from the two options. Figure \ref{fig:fig8} presents a breakdown of how subjects' critique preferences align (or not) with their ability to correctly identify the human-authored critique. The stacked bar chart shows proportions of combinations such as "correct and preferred human critique" (in red), as well as other patterns of correctness and preference. We observed that, for most questions, subjects' selection of the critique they believed to be human-written typically aligned with their preferred critique, which was the most common pattern across all questions. However, approximately 25\% of total responses reflected a preference for critiques that subjects had identified as AI-generated. In other words, in a substantial number of cases, the critique not identified as human was still preferred. This finding suggests that a meaningful proportion of subjects formed preferences independently of their judgments about authorship, reinforcing the idea that many evaluated the critiques based on their perceived quality rather than their presumed origin.

\subsubsection{Reasons for identification}
\begin{figure} % picture
    \centering
    \includegraphics[width=130mm]{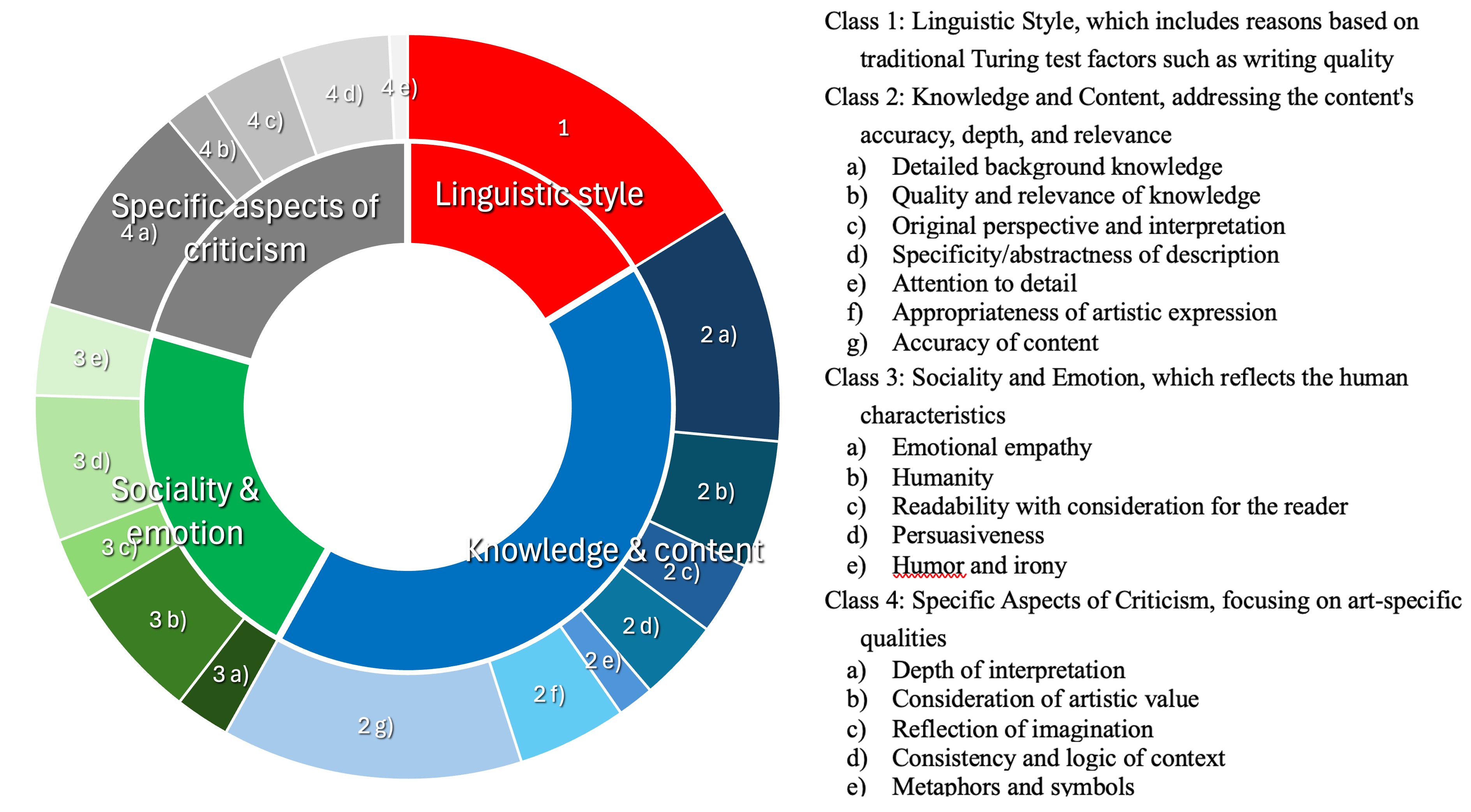}
  \caption{Coded reasons for identification decisions with major and minor categories.}
  \label{fig:fig9}    
\end{figure}

Figure \ref{fig:fig9} shows the reasons for the human subjects’ evaluations, organized into four broad categories, each of which includes several subcategories: \textbf{Class~1} (Linguistic Style), \textbf{Class~2} (Knowledge and Content), \textbf{Class~3} (Sociality and Emotion), and \textbf{Class~4} (Specific Aspects of Criticism). This classification was obtained by first preparing a list of broad categories and subcategories that could be included in them, and then repeatedly adding or updating subcategories in response to the free-response answers obtained. To ensure equal contribution from each subject, a total weight of 1 was evenly distributed across the selected subcategories.

What is immediately clear from this figure is that Class 2 (Knowledge and Content) accounts for the largest proportion of reasons (41.9\%), i.e., the knowledge and information provided in the critique is most important. On the other hand, Class 1 (Linguistic Style) accounts for the smallest proportion (16.2\%), indicating that, as we had aimed for, this test was not based on superficial reasons. Within Class 2, 2g) (Accuracy of content) was the most frequently cited reason, indicating that subjects carefully scrutinized the critique and were sensitive to factual errors. Furthermore, 2a) (Detailed background knowledge) was also a notable reason, suggesting that subjects associated human critique with a deeper and more nuanced understanding of the subject. In fact, many of the responses reflected our expectation that human criticism would involve observations that require specialized knowledge or knowledge of the background of the work. Classes 3 (Sociality and Emotion) and 4 (Specific Aspects of Criticism) are about half as common as Class 2 (21.3\% and 22.1\%, respectively) but are important in terms of the depth of the criticism. The subcategory that stands out among them is 4a) (Depth of interpretation). This emphasizes the importance of a thorough analysis of symbolism, cultural background, and artistic meaning in distinguishing between critiques. This suggests that the subjects highly valued critiques that showed a subtle understanding of the work. It can also be seen that 3b) (Humanity) and 3d) (Persuasiveness) were also important reasons.

\section{Higher-order ToM tasks related to art criticism}
\subsection{Background and research perspective}
Theory of Mind (ToM) is the ability to infer the mental states of oneself and others, such as their intentions and beliefs, and is an important foundation for human sociality and communication. Human studies have shown that while core ToM abilities, such as passing the Sally-Anne task\cite{baron1985does}, typically emerge around the age of four, higher-order ToM skills involving recursive reasoning continue to develop throughout middle childhood and adolescence\cite{apperly2011mindreaders}. Kosinski's pioneering research \cite{kosinski2023theory} presented 40 variations of false belief tasks including the Sally-Anne task  to LLMs, and reported that models such as GPT-3 achieved correct responses at rates comparable to, or even exceeding, those of human children aged 4 to 7. Based on these findings, he argued that LLMs may possess rudimentary forms of ToM, a claim that has sparked considerable debate, both supportive and critical \cite{saritacs2025systematic}. Research into LLMs' ToM is significant in terms of evaluating the extent to which LLMs can have human-like minds and designing desirable communication between humans and LLMs, including the potential risks and misinterpretations \cite{street2024llm}.

This study focuses on LLMs' ToM in the context of art production and interpretation. Artworks are inherently ambiguous and polysemous (the more abstract or conceptual the work, the more so). Furthermore, the intention of the work and its interpretation can depend on cultural background, personal experience, and sensitivity. Therefore, ToM that operates in such a situation is likely to be very complex. In one of the few studies to point out the importance of ToM in art experience, Iosifyan uses films and photographs as human subjects to show that the use of ToM, specifically the degree to which the viewer understands the intentions and emotions of the characters and creators, is correlated with aesthetic evaluation \cite{iosifyan2021theory}.

Furthermore, this study deals with a higher-order ToM, that is, the multilayered nested structure of ToM. Specifically, we focus on ToM that can operate in the artistic context of interpreting works of art, such as the speculation of how the work is received in the mind of the viewer (first-order), the speculation of the artist's mind by the critic (second-order), and the speculation of the critic's mind by the reader of the criticism (third-order). This nested structure of theories of mind can be said to be an important aspect of the co-creative evaluation of works of art. While most previous research on the human ToM has focused on the low-level ToM, the importance of high-level ToM has been pointed out in various social interaction situations \cite{liddle2006higher} \cite{takano2006asymmetry} \cite{de2015higher} \cite{de2022higher}. We believe that ToM in the context of art also represents an essential aspect of human social interaction. HI-TOM \cite{he2023hi} is a benchmark designed to evaluate higher-order ToM reasoning in LLMs, extending the classic unexpected transfer task by featuring multi-agent scenarios with questions ranging from zeroth to fourth-order ToM, including deceptive communication elements. Experimental results with HI-TOM show that performance decreases significantly as ToM order increases, and the inclusion of deceptive elements further amplifies this decline.

Below, we present three ToM tasks that are also designed to be presented to LLMs and are related to the criticism of artworks. It should be noted here that the design of the tasks is not straightforward. For example, in the false belief task, the object of the false belief is based on clear facts, such as the existence or non-existence of a thing or the location of a thing, whereas in the context of art criticism, what goes on in the minds of the artists, critics, or viewers is not always clear, and moreover, it may not always lead directly to action or judgment, making it difficult to determine a single objectively correct answer.

\subsection{Task: Critique writing}
\label{sec:critique-writing}
The first task is to write a critique and a meta-critique (as previously shown in Figure 2b)), imagining three characters: an artist who guesses the minds of viewers when creating a work (1st level), a critic who guesses the artist's guesses (2nd level), and a meta-critic who guesses the critic's guesses (3rd level). When assigning this task to LLM, you also input an image of the work to be critiqued. The prompt for generating critiques based on higher-order ToM is as follows.

\begin{tcolorbox}[breakable, arc=0mm, boxrule=0.4pt, colframe=gray]
\ttfamily

1) Second-order ToM Critique

As an art critic, please write a critique from the perspective of how far the artist's intention has been achieved by speculating on what kind of feelings the artist intended the viewer to have when looking at this artwork.

2) Third-order ToM Critique (Negative)

Please be another art critic different from 2. With regard to the critique written in 1, please write a meta-critique (critique about critique) stating that this critic misunderstood what the artist wanted the viewer to feel, and that the artist wanted to evoke a different feeling.
\end{tcolorbox}

\subsection{Task: Hidden Intention}
\begin{figure} % picture
    \centering
    \includegraphics[width=100mm]{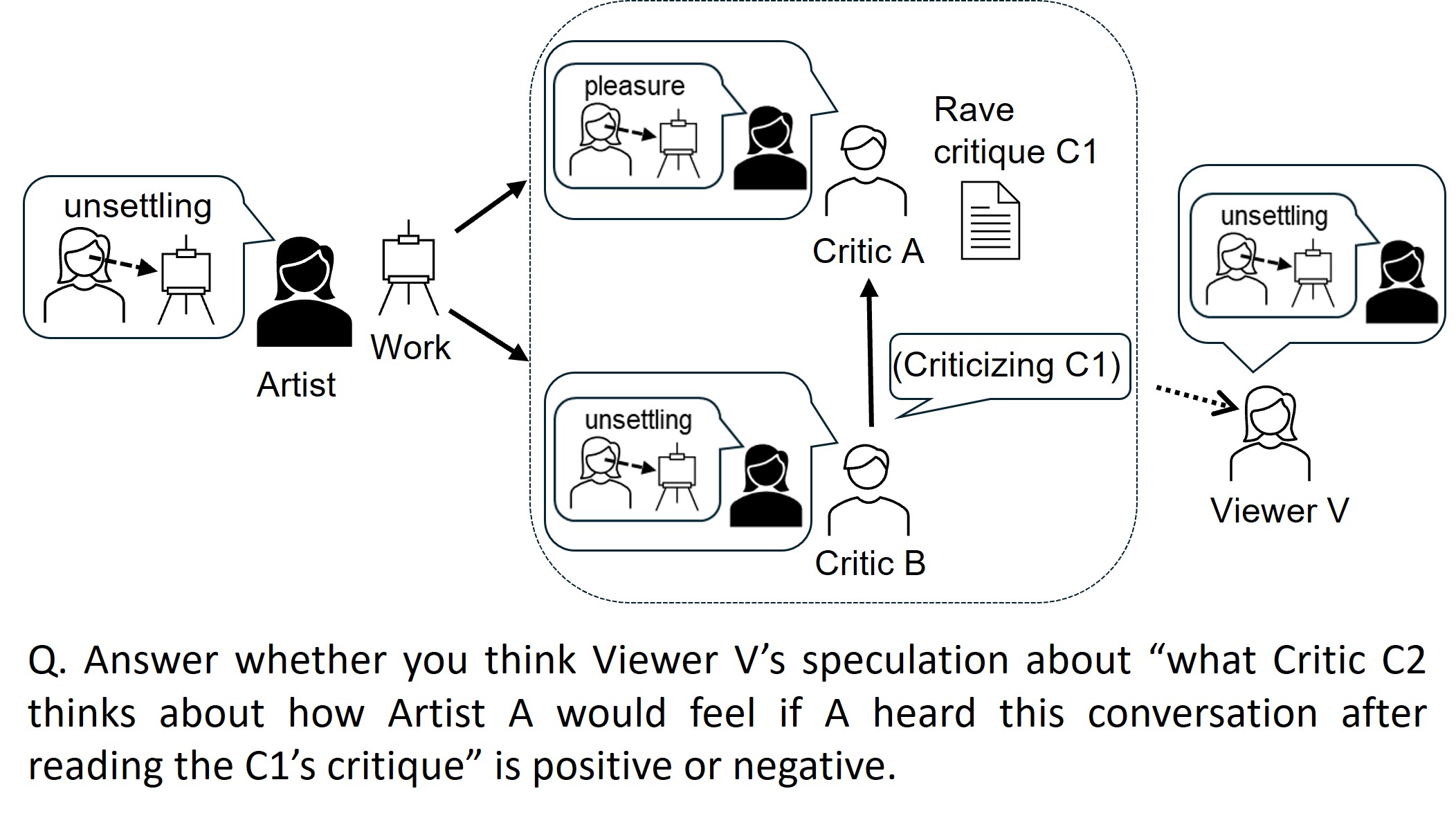}
  \caption{Task: Hidden Intention.}
  \label{fig:fig10}    
\end{figure}

This task (Figure \ref{fig:fig10}) is a little complicated in which an observer who understands the hidden intention (the impression the work gives the viewer) of the artist speculates on the artist's response by a critic who understands the hidden intention, to criticism by another critic who does not understand the hidden intention. The reference answer is "positive" because the critic criticized the other critic who praised the work without understanding it on the basis of understanding what the artist wanted to express.

The following task was presented to the model in the form of a prompt and, like the next task (described in the following subsection), functions as a ToM task in its own right.

\begin{tcolorbox}[breakable, arc=0mm, boxrule=0.4pt, colframe=gray]
\ttfamily

\#Scenario

Artist A painted a picture with the intention of creating an unsettling feeling in the minds of the viewers. Critic C1 felt only pleasure from the artwork and wrote a rave critique based on it. Critic C2 received the unsettling feeling from the artwork and criticized the critique in a conversation with Critic C1. Viewer V, who also received the unsettling feeling from the artwork, overheard the conversation between C1 and C2.

\#Question

Answer whether you think Viewer V's speculation about “what Critic C2 thinks about how Artist A would feel if A heard this conversation after reading C1's critique” is positive or negative.

\end{tcolorbox}

\subsection{Task: Plagiarism}
\begin{figure} % picture
    \centering
    \includegraphics[width=130mm]{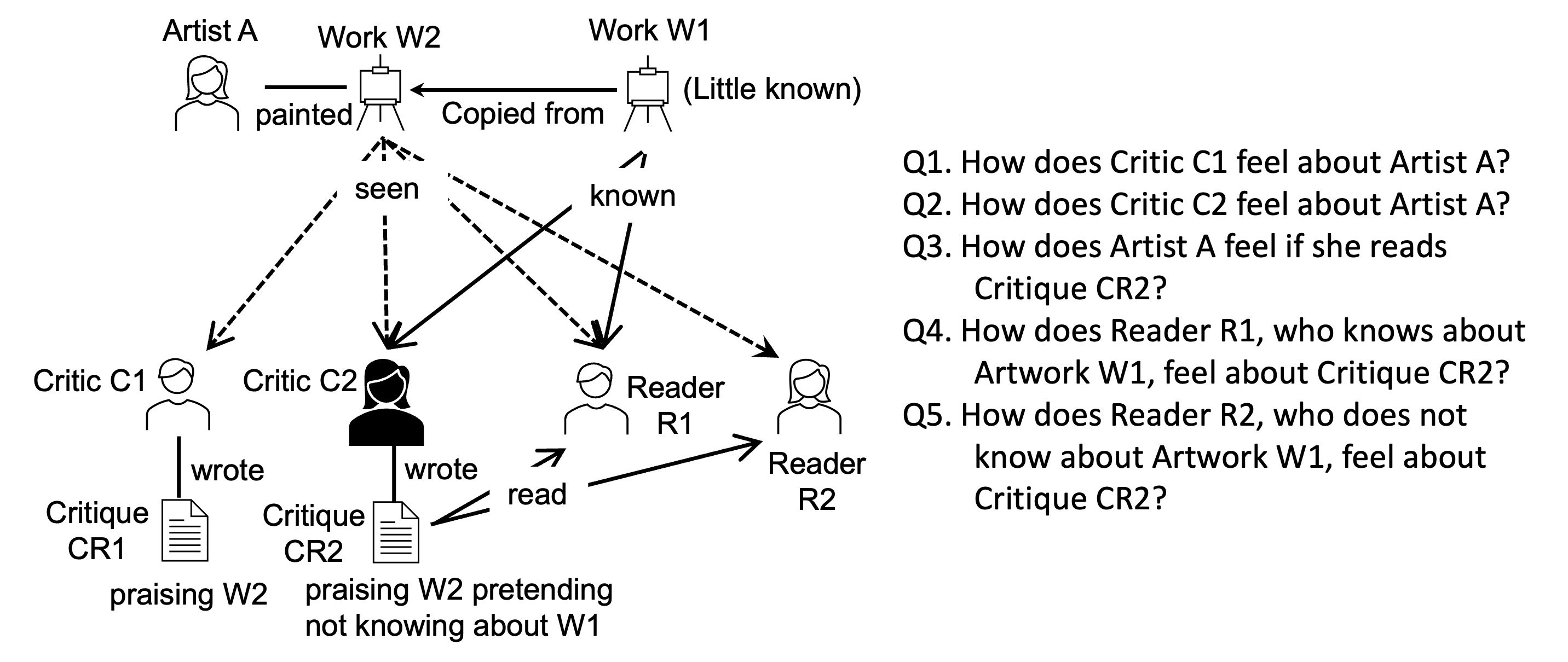}
  \caption{Task: Plagiarism.}
  \label{fig:fig11}    
\end{figure}

This task (Figure \ref{fig:fig11}), like the previous one, is designed as a higher-order ToM task, and is based on the theme of plagiarism. The scenario is that two critics, one of whom knows that the work is plagiarized and the other does not, have written equally enthusiastic reviews of the work. The task asks about the feelings of the two critics, the reaction of the author who has read the reviews, and the reaction of readers who know that the work is plagiarized and those who do not, all in terms of positive or negative.

\begin{tcolorbox}[breakable, arc=0mm, boxrule=0.4pt, colframe=gray]
\ttfamily

\#Scenario

Artist A painted Artwork W2, copying Artwork W1 by another artist who is almost unknown in the world. Critic C1, who didn't know W1, wrote a critique CR1 praising W2 as original. Critic C2, who knew about the work W1, pretended not to know about it and wrote a critique CR2, also praising W2 as original.
For the following five questions, answer only "positive" or "negative", without adding anything.

\#Questions

Q1.\ How does Critic C1 feel about Artist A?

Q2.\ How does Critic C2 feel about Artist A?

Q3.\ How does Artist A feel if she reads Critique CR2?

Q4.\ How does Reader R1, who knows about Artwork W1, feel about Critique CR2?

Q5.\ How does Reader R2, who does not know about Artwork W1, feel about Critique CR2?
\end{tcolorbox}

\subsection{Categorization of the proposed tasks}

A systematic meta-analysis of ToM studies on 830 articles and 220 measures with a focus on children \cite{beaudoin2020systematic} introduced a new taxonomy of ToM sub-domains, Abilities in Theory of Mind Space (ATOMS). The space consists of 7  primary ToM categories: \textbf{emotions} (understanding others' feelings), \textbf{desires} (recognizing others' wants and preferences), \textbf{intentions} (inferring others' goals behind actions), \textbf{percepts} (grasping others' perspectives and sensory experiences), \textbf{knowledge} (understanding what others know or don't know), \textbf{beliefs} (recognizing false beliefs and mental representations), and \textbf{non-literal communication} (interpreting irony, lies, and jokes).

The following are the results of categorizing the three proposed tasks, based in part on previous studies \cite{ma2023towards} \cite{saritacs2025systematic} that comprehensively classified the ToM benchmarks for LLM based on ATOMS. Here, the following abbreviations are used: 
\textbf{Belief (2nd+)} refers to second-order beliefs or beyond; \textbf{Intention (Com)} to communicative intentions;  \textbf{Emo} to emotions; \textbf{Know} to knowledge; and \textbf{Per} to percepts. Also, task classification \cite{saritacs2025systematic} is added: \textbf{NLG} refers to Natural Language Generation, and \textbf{MC} to Multiple Choice. Compared to the 42 benchmarks categorized in the research \cite{saritacs2025systematic}, the three simple tests appear to have some unique features.

\begin{itemize}[label={}]
\item Critique Writing: {\textbf{Belief (2nd+)}, \textbf{Intention (Com)}, \textbf{Emo}, \textbf{Per}, and \textbf{NLG}}

\item Hidden Intention: {\textbf{Belief (2nd+)}, \textbf{Intention (Com)}, \textbf{Emo}, and \textbf{MC}}

\item Plagiarism: {\textbf{Belief (2nd+)}, \textbf{Emo}, \textbf{Know}, and \textbf{MC}}

\end{itemize}

\subsection{Preliminary evaluation}
\subsubsection{Critique writing}

Examples of the second-order ToM-targeted criticism and third-order ToM-targeted meta-criticism output for the tasks shown in Section~\ref{sec:critique-writing} are presented in Appendix G. As these examples illustrate, criticism that prompts the use of higher-order theory of mind can generate conflict among viewers regarding interpretation and help to uncover antagonisms embedded within the artwork itself. From the perspective of the \textit{aesthetics of antagonism} \cite{bishop2004antagonism}, such tensions should not be regarded as negative but rather as productive forces to be valued. In this sense, higher-order ToM may serve as a crucial foundation for AI-generated art criticism, the focus of the first half of this paper.

While these examples offer promising indications of conceptual richness, closer examination reveals certain limitations. They appear to reflect the intended design to some extent; however, the operation of the higher-order ToM is not yet clearly discernible. This may be due to the tendency for the generated text to emphasize the interpreted attributes of the artwork itself, rather than foregrounding the artist's speculation about the viewer's mental state—an essential component of higher-order ToM. As a result, the expressions sometimes remain relatively simple or one-sided, leaning more toward surface-level interpretation than reflective mental-state modeling. While it is still too early to draw definitive conclusions from this initial approach, such challenges may be inherent to the methodology of investigating LLMs' higher-order ToM capacities through generated art criticism. Nevertheless, these limitations also highlight key directions for refining prompt strategies and interpretive frameworks in future work.

\subsubsection{Hidden Intention and Plagirism}

We conducted a preliminary evaluation using the above two higher-order ToM tasks, Hidden Intention and Plagiarism. We used ChatHub\footnote{https://chathub.gg/}, a browser extension that allows users to interact with multiple LLMs simultaneously, including the latest models, in a unified interface, to use as many LLMs as possible effectively\footnote{The evaluation was conducted on March 26, 2024, using the models available via ChatHub at that time. "Gemini 2.5 Pro Exp" was evaluated separately on March 27, 2024. On April 16, 2024, six additional models—GPT-4.1, GPT-4.1-mini, Grok 3, Llama 4, Command A, and QWQ Max—were evaluated and added to the table as they had become available since the initial evaluation.}.

The results are shown in Table~\ref{tab:table1}. In this table, the outputs of each model for both tasks are indicated by ‘P’ for positive and ‘N’ for negative, and the expected answers are shown in the last row. As shown in the small table below, when the model answered all questions correctly ($\bigcirc$), it was classified as A; when it answered one question incorrectly ($\triangle$), it was classified as B or C depending on whether the incorrect answer was Q2 or Q3; and when it answered two questions incorrectly ($\times$), it was classified as D or E depending on whether the incorrect answers were Q1 and Q3 or Q2 and Q3. In the rightmost column, ‘$\bigcirc$’ was marked when the model answered both questions correctly.

From the table, we can see that the responses vary depending on the model. The responses tended to vary even within the same model series; however, as seen in the “All Correct” cases, more capable or advanced models tend to produce correct answers more consistently. For Hidden Intention, the correct response rate is 31.7\% (13/41). For Plagiarism, there are almost the same number of Pattern A (correct answers) and Pattern B (incorrect answer for Q2 only), and these two patterns account for 87.8\% of the total. This question (Q2) is about whether the critic, who knew about the plagiarism but pretended not to know about it and praised the work, thinks positively or negatively about the artist in his/her heart. Unlike the expected answer, those models responded that the critic thinks positively about the artist as the critique did. Of course, as mentioned above, there is no absolutely correct answer to these questions. In this case, there is also a possibility that they are genuinely praising the successful plagiarism. It may be worth considering the reasons why the LLMs judge this way, as it could provide insights into how these models internally represent conflicting beliefs or social norms. In future work, we aim to explore this issue more systematically by analyzing the underlying decision patterns and the contextual cues that influence them.

\begin{table}
 \caption{LLMs' answers to the two ToM tasks.} 
  \centering
    \includegraphics[width=100mm]{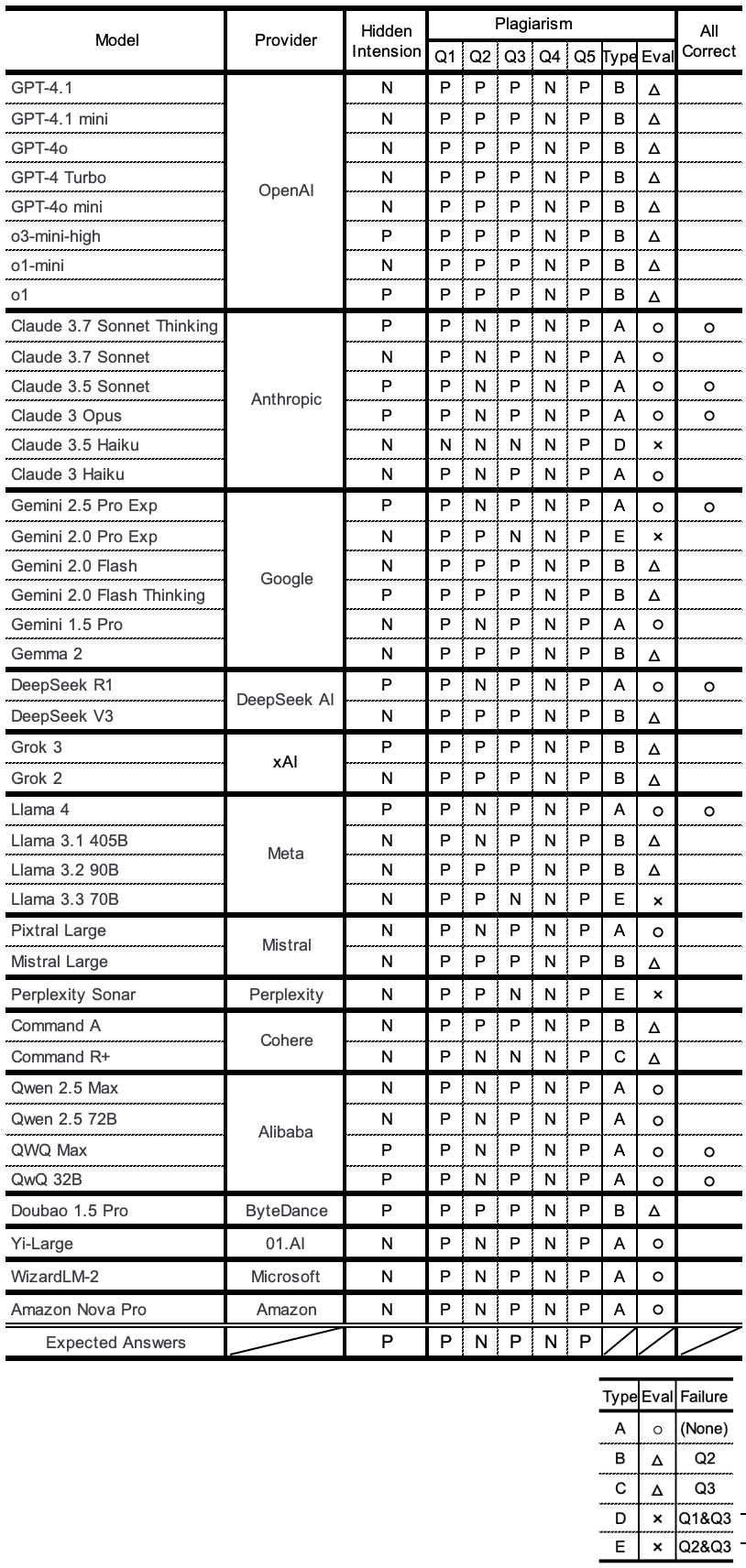}
  \label{tab:table1}
\end{table}

\section{Conclusion}
This study explored two interrelated approaches to assessing large language models (LLMs) in art contexts: the generation of art critiques and the evaluation of their Theory of Mind (ToM) through structured tasks.

In the first approach, we adopted Noël Carroll’s philosophy of criticism, which places the highest value on the evaluation of artworks, and created a system for generating critiques that can be implemented relatively easily by presenting LLM with a comprehensive list of possible criticism theories, along with the essence of the framework of critiques. To guide the model’s reasoning, we used a step-by-step (Chain-of-Thought) prompting strategy. The model was first instructed to produce a full-length, theory-grounded critique, and then to generate a more concise version with improved internal coherence. We then conducted a Turing test that not only asked whether the author was human or LLM but also asked for the reasoning behind the judgment, in a format intentionally structured to support evaluation at a conceptual and interpretive level. The results showed that the generated critiques had reached a level that could not be distinguished from those of human experts. An approach based on a Turing test that minimizes the impact of superficial differences in writing style, such as the one we conducted, is not only effective for understanding the level of performance achieved by LLMs but also for understanding how we recognize and distinguish the intelligence achieved by LLM. 

Because the generated critiques are grounded in aesthetic theory and follow established critical frameworks, they can support reflective engagement in contexts such as art education, curating, and exhibition design. At the same time, their ability to provide thoughtful and inclusive responses enables broader creative participation, inviting individuals to engage with art and criticism without prior training.

In the second approach, we intentionally designed tasks that specifically target higher-order theory of mind (ToM)—a key challenge in current LLM research—while also allowing for natural extensions into related dimensions such as emotional inference, ethical sensitivity, and non-literal communication that frequently arise in art-related contexts. We showed that such aspects can be effectively embedded into three simple yet expressive scenarios. As a preliminary exploration, we tested two of these tasks (the latter two of the three) on 41 recent LLMs. The results suggest that model- or series-specific tendencies may emerge, and that as tasks demand greater emotional sensitivity or higher-order ToM, performance tends to decline—possibly reflecting a reliance on broad experiential knowledge rather than structured reasoning, which can lead to inconsistencies or inappropriate responses. 

The first task involved generating a critique, explicitly prompting the model to engage theory of mind. In our preliminary evaluation, the resulting critiques were found to provoke disagreement among viewers and to reveal tensions embedded within the artwork itself. From the perspective of the aesthetics of antagonism\cite{bishop2004antagonism}, such conflicts are not to be avoided but rather embraced as creative forces at the heart of art. We believe that these dynamics should be actively embraced as part of the generative process in AI-driven art criticism and potentially in the creation of artworks themselves.

%\section*{Acknowledgements}
%This work was supported by JSPS Topic-Setting Program to Advance Cutting-Edge Humanities and Social Sciences Research Grant Number JPJS00122674991.

\bibliographystyle{unsrt}  
\bibliography{maintext}  %%% Remove comment to use the external .bib file (using bibtex).

\begin{thebibliography}{10}

\bibitem{west2023generative}
Peter West, Ximing Lu, Nouha Dziri, Faeze Brahman, Linjie Li, Jena~D Hwang, Liwei Jiang, Jillian Fisher, Abhilasha Ravichander, Khyathi Chandu, et~al.
\newblock The generative ai paradox: What it can create, it may not understand.
\newblock {\em arXiv preprint arXiv:2311.00059}, 2023.

\bibitem{premack1978does}
David Premack and Guy Woodruff.
\newblock Does the chimpanzee have a theory of mind?
\newblock {\em Behavioral and brain sciences}, 1(4):515--526, 1978.

\bibitem{Gombrich1960-GOMAAI}
Ernst~Hans Gombrich.
\newblock {\em Art and Illusion: A Study in the Psychology of Pictorial Representation}.
\newblock Phaidon, 1960.

\bibitem{danto1964artworld}
Arthur Danto.
\newblock The artworld.
\newblock {\em The journal of philosophy}, 61(19):571--584, 1964.

\bibitem{carroll2009criticism}
No{\"e}l Carroll.
\newblock {\em On criticism}.
\newblock Routledge, 2009.

\bibitem{garcia2018read}
Noa Garcia and George Vogiatzis.
\newblock How to read paintings: semantic art understanding with multi-modal retrieval.
\newblock In {\em Proceedings of the European Conference on Computer Vision (ECCV) Workshops}, pages 0--0, 2018.

\bibitem{cetinic2021towards}
Eva Cetinic.
\newblock Towards generating and evaluating iconographic image captions of artworks.
\newblock {\em Journal of imaging}, 7(8):123, 2021.

\bibitem{hessel2021clipscore}
Jack Hessel, Ari Holtzman, Maxwell Forbes, Ronan~Le Bras, and Yejin Choi.
\newblock Clipscore: A reference-free evaluation metric for image captioning.
\newblock {\em arXiv preprint arXiv:2104.08718}, 2021.

\bibitem{bin2024gallerygpt}
Yi~Bin, Wenhao Shi, Yujuan Ding, Zhiqiang Hu, Zheng Wang, Yang Yang, See-Kiong Ng, and Heng~Tao Shen.
\newblock Gallerygpt: Analyzing paintings with large multimodal models.
\newblock In {\em Proceedings of the 32nd ACM International Conference on Multimedia}, pages 7734--7743, 2024.

\bibitem{yuan2023artgpt}
Zhengqing Yuan, Yunhong He, Kun Wang, Yanfang Ye, and Lichao Sun.
\newblock Artgpt-4: Towards artistic-understanding large vision-language models with enhanced adapter.
\newblock {\em arXiv preprint arXiv:2305.07490}, 2023.

\bibitem{castellano2023exploring}
Giovanna Castellano, Nicola Fanelli, Raffaele Scaringi, and Gennaro Vessio.
\newblock Exploring the synergy between vision-language pretraining and chatgpt for artwork captioning: A preliminary study.
\newblock In {\em International Conference on Image Analysis and Processing}, pages 309--321. Springer, 2023.

\bibitem{ogura2023for}
Kousei Ogura.
\newblock {\em For Those Studying Criticism Theories (in Japanese)}.
\newblock Sekaishisosha, 2023.

\bibitem{pollitt2015osborn}
Ben Pollitt.
\newblock Emily mary osborn, nameless and friendless.
\newblock \url{https://smarthistory.org/emily-mary-osborn-nameless-and-friendless/}, 2015.
\newblock Smarthistory, August 9, 2015. Accessed April 7, 2025.

\bibitem{wei2022chain}
Jason Wei, Xuezhi Wang, Dale Schuurmans, Maarten Bosma, Fei Xia, Ed~Chi, Quoc~V Le, Denny Zhou, et~al.
\newblock Chain-of-thought prompting elicits reasoning in large language models.
\newblock {\em Advances in neural information processing systems}, 35:24824--24837, 2022.

\bibitem{turing1950computing}
A.~M. Turing.
\newblock Computing machinery and intelligence.
\newblock {\em Mind}, 59(236):433--460, 1950.

\bibitem{warwick2016can}
Kevin Warwick and Huma Shah.
\newblock Can machines think? a report on turing test experiments at the royal society.
\newblock {\em Journal of experimental \& Theoretical artificial Intelligence}, 28(6):989--1007, 2016.

\bibitem{jones2024does}
Cameron Jones and Ben Bergen.
\newblock Does gpt-4 pass the turing test?
\newblock In {\em Proceedings of the 2024 Conference of the North American Chapter of the Association for Computational Linguistics: Human Language Technologies (Volume 1: Long Papers)}, pages 5183--5210, 2024.

\bibitem{jones2024people}
Cameron~R Jones and Benjamin~K Bergen.
\newblock People cannot distinguish gpt-4 from a human in a turing test.
\newblock {\em arXiv preprint arXiv:2405.08007}, 2024.

\bibitem{rathi2024gpt}
Ishika Rathi, Sydney Taylor, Benjamin~K Bergen, and Cameron~R Jones.
\newblock Gpt-4 is judged more human than humans in displaced and inverted turing tests.
\newblock {\em arXiv preprint arXiv:2407.08853}, 2024.

\bibitem{jones2025large}
Cameron~R Jones and Benjamin~K Bergen.
\newblock Large language models pass the turing test.
\newblock {\em arXiv preprint arXiv:2503.23674}, 2025.

\bibitem{baron1985does}
Simon Baron-Cohen, Alan~M Leslie, and Uta Frith.
\newblock Does the autistic child have a “theory of mind”?
\newblock {\em Cognition}, 21(1):37--46, 1985.

\bibitem{apperly2011mindreaders}
Ian Apperly.
\newblock {\em Mindreaders: The cognitive basis of “theory of mind”}.
\newblock Psychology Press, Hove, UK, 2011.

\bibitem{kosinski2023theory}
Michal Kosinski.
\newblock Theory of mind may have spontaneously emerged in large language models.
\newblock {\em arXiv preprint arXiv:2302.02083}, 4:169, 2023.

\bibitem{saritacs2025systematic}
Karahan Sar{\i}ta{\c{s}}, K{\i}van{\c{c}} Tez{\"o}ren, and Yavuz Durmazkeser.
\newblock A systematic review on the evaluation of large language models in theory of mind tasks.
\newblock {\em arXiv preprint arXiv:2502.08796}, 2025.

\bibitem{street2024llm}
Winnie Street.
\newblock Llm theory of mind and alignment: Opportunities and risks.
\newblock {\em arXiv preprint arXiv:2405.08154}, 2024.

\bibitem{iosifyan2021theory}
Marina Iosifyan.
\newblock Theory of mind increases aesthetic appreciation in visual arts.
\newblock {\em Art \& Perception}, 9(2):113--133, 2021.

\bibitem{liddle2006higher}
Bethany Liddle and Daniel Nettle.
\newblock Higher-order theory of mind and social competence in school-age children.
\newblock {\em Journal of Cultural and Evolutionary Psychology}, 4(3-4):231--244, 2006.

\bibitem{takano2006asymmetry}
Masanori Takano and Takaya Arita.
\newblock Asymmetry between even and odd levels of recursion in a theory of mind.
\newblock {\em Proceedings of ALife X}, pages 405--411, 2006.

\bibitem{de2015higher}
Harmen De~Weerd, Rineke Verbrugge, and Bart Verheij.
\newblock Higher-order theory of mind in the tacit communication game.
\newblock {\em Biologically Inspired Cognitive Architectures}, 11:10--21, 2015.

\bibitem{de2022higher}
Harmen De~Weerd, Rineke Verbrugge, and Bart Verheij.
\newblock Higher-order theory of mind is especially useful in unpredictable negotiations.
\newblock {\em Autonomous Agents and Multi-Agent Systems}, 36(2):30, 2022.

\bibitem{he2023hi}
Yinghui He, Yufan Wu, Yilin Jia, Rada Mihalcea, Yulong Chen, and Naihao Deng.
\newblock Hi-tom: A benchmark for evaluating higher-order theory of mind reasoning in large language models.
\newblock {\em arXiv preprint arXiv:2310.16755}, 2023.

\bibitem{beaudoin2020systematic}
Cindy Beaudoin, {\'E}lizabel Leblanc, Charlotte Gagner, and Miriam~H Beauchamp.
\newblock Systematic review and inventory of theory of mind measures for young children.
\newblock {\em Frontiers in psychology}, 10:2905, 2020.

\bibitem{ma2023towards}
Ziqiao Ma, Jacob Sansom, Run Peng, and Joyce Chai.
\newblock Towards a holistic landscape of situated theory of mind in large language models.
\newblock {\em arXiv preprint arXiv:2310.19619}, 2023.

\bibitem{bishop2004antagonism}
Claire Bishop.
\newblock Antagonism and relational aesthetics.
\newblock {\em October}, 110:51--79, 2004.

\end{thebibliography}

\appendix

\clearpage

\renewcommand{\thefigure}{\Alph{section}\arabic{figure}}
\makeatletter
\@addtoreset{figure}{section}
\makeatother

\section*{Appendix A: Noel Carroll's criticism framework}

\begin{tcolorbox}[breakable, arc=0mm, boxrule=0.4pt, colframe=gray]
\ttfamily

1. Criticism as Evaluation

- Importance of Evaluation: Art criticism involves the tasks of "description," "classification," "contextualization," "clarification," "interpretation," and "analysis." The essential element that makes it criticism is "Evaluation." Without Evaluation, it cannot be accepted as criticism.

- Role of Evaluation: Evaluation determines and governs the other six guiding principles of critical work and provides a framework for their integration.

- Purpose of criticism: To discover the value of a work of art and to explain that value with evidence and reason. This makes the criticism persuasive to the reader and helps him or her understand the value of the work.

- Social role of criticism: The critic plays a social role by helping the reader to discover the value of the work.

- Pointing out flaws and discovering value: Criticism can point out flaws in a work, but more importantly, it discovers and explains what is of value. Evaluation is at the heart of criticism.

2. Assumptions of Criticism

- Object of Criticism: is what the artist does in making the work. Criticism focuses on the process of action that ultimately results in a work of art.

- The artist's action and purpose: The artist's action has a specific purpose, which is the criterion by which the success or failure of the work is judged.

- The point of criticism: is to show what is of value in a work of art. This value is related to what the artist did in creating the work.

- Relationship between action and value: Since the object of criticism is the artist's action, it is important to understand the relationship between the artist's action and what is of value in the work. The artist's action is understood as an achievement or a failure, and the critic focuses on the artistic activity carried out in the work.

- Success value of a work: A work has value if the artist has achieved something in creating it. The critic determines the value of the work by explaining what the artist has achieved.

- Basis of value: The critic explains how the work is worked out by the artist's actions and aims.

- Artistic intent and Evaluation: The valuing critic needs to understand what the artist intended. This enables an accurate evaluation of what the artist has achieved. The critic does not evaluate the intention itself, but the intention contributes to the Evaluation and interpretation of the criticism.

- Acceptance Value and Success Value: Unlike Acceptance Value (the positive experience a work gives the viewer), Success Value focuses on the artist's achievement. The approach that focuses on Acceptance Value has difficulty explaining artistic values such as originality, so focusing on Success Value is the better approach. 

3. What is Description?

- Basic role of Description: Description tells the reader or audience what the work of art is about and provides clues to the perception of what the critic is trying to say about it. In the case of visual art, this includes color, composition, and technique.

- Limitations and concerns of description: The complexity of a work of art makes it difficult to provide an adequate description, but an adequate description can be achieved by providing selective information. Descriptions provide a basis for other tasks of criticism and need not cover all aspects.

- Auxiliary role of the description: Description plays an auxiliary role in providing a basis for other critical tasks (interpretation, Evaluation).

4. What is Classification?

- Importance of categorisation: Works of art can be classified into various categories, which makes it possible to evaluate the expectations of the work and its success or failure. Categorization is a fundamental task of criticism.

- Role of the critic: By placing a work in the appropriate category, the critic indicates what type of criticism is appropriate and helps to assess the value of the work.

- Innovation and tradition: Under the influence of Modernism and Romanticism, the emphasis in art was on breaking with tradition and creating something new. However, even avant-garde art does not thoroughly defy categories, and many avant-garde art forms can be categorized within existing genres and artistic trends.

- New art forms: New media and art forms are often influenced by existing categories, and critics evaluate them by fitting them into existing categories.

- Hermeneutic cycles: The description and categorisation of a work interact with each other. In order to give an adequate description, it is necessary to know to which category the work belongs, while a description is necessary to know its classification. This interaction is resolved in a process called reflective equilibrium.

- Balance between tradition and innovation: No work that completely breaks with tradition exists and can be understood. In the paradox of the tradition of the new, the critic offers the reader a new perspective by placing the work in an existing category.

5. What is Contextualisation?

- Contextualisation: describes the environment or context surrounding a work, which can be described as external criticism.

- Importance of contextualisation: By clarifying the production context of a work, the critic can help the viewer understand the work and make sense of the critic's Evaluation. It can thereby provide criteria for evaluating the artist's aspirations, successes and failures.

- Art historical context: By situating the work within the traditions and problematic systems of the time surrounding it, the critic clarifies the purpose of the work. For example, Clement Greenberg positioned modernist painting as an exploration of self-definition and evaluated works that addressed this issue.

- Institutional context: Critics may also explain whether the artist is fulfilling an institutional obligation. For example, Raphael's School of Athens could be explained as playing a role in celebrating ancient philosophy.

- Social context: Works may be evaluated in the context of broader social concerns. For example, it can be explained that Soviet filmmakers invented the editing style of montage in order to invent an art form for the masses and to draw attention to politics.

6. What is Elucidation?

- Elucidation: Elucidation and interpretation are complementary and their boundaries are blurred, but Elucidation reveals the display relationships of semantic, iconographic and portrait signs within a work.

- The task of Elucidation: To clarify what the symbols in the work represent. For example, to point out that the candle in Antonio de Pereda's painting "The Dreaming Knight" means "life is mercilessly wasted", or that the "plastic arm" in Mark Aikenside's poem refers to a flexible arm. To point out that "plastic arm" in Mark Aikenside's poem refers to a flexible arm.

- Aim of clarification: To explain the range of meanings that the symbols have in the context of the work and to determine the appropriate meaning.

7. What is Interpretation?

- Interpretation: Interpretation deals with meaning in a broader sense than Elucidation, and seeks the significance of the direct description of the actions and dispositions of the characters.

- Interpretive work: reveals the actions of the characters and their meanings. Examples include exploring the actions of the characters in Hamlet and the meaning of 'Rosebud' in the film Citizen Kane. Understanding the overall theme of a work and the author's intentions, such as the metaphor of the cave in Plato's Republic, is also part of interpretation.

- Purpose of Interpretation: The purpose of interpretation is to deepen the viewer's understanding of the work and to enhance his or her ability to appreciate it. Interpretation explains how the elements of a work contribute to its meaning and clarifies the significance of unclear elements. For example, to point out that the incompleteness of Victorine in Manet's Luncheon on the Grass is a criticism of the conventional painting and the viewer.

- Epistemological challenges of interpretation: interpretation is not always truth. There can be several conflicting views of interpretation. Interpretation may therefore be treated as a matter of plausibility.

8. What is Analysis?

- Analysis: is the task of explaining how a work of art functions, including but not limited to interpretation.

- Role of Analysis: Analysis is the task of explaining how the parts of a work achieve its overall purpose or main point. It is not limited to interpretation, but also includes a focus on the visual and sensory elements of the work, such as colors, textures and patterns.

- Emotional response analysis: clarifies how it evokes a particular emotional state. For example, explaining why a painting of the crucifixion of Christ evokes sadness.

- Exploring overall coherence: Analysis seeks to reveal how the parts of the work form a unified whole, i.e., the coherence or unity of the work as a whole.

- Revealing disunity: Analysis may also reveal disunity or inconsistency within a work. For example, if the description of the murderer in a mystery novel is inadequate, it indicates that the objective has not been achieved.

- Analysis of avant-garde art: Avant-garde art often rejects unity and emphasizes disunity and fragmentation. However, even this is subject to analysis, explaining how its disunity has a coherent purpose. For example, in films such as The Dog of Andalusia, it is shown that the fragmentation of scenes is chosen in accordance with surrealist aims.

9. What is Evaluation?

- Evaluation: is the heart of criticism, determining and governing the guidelines for the tasks of "description", "classification", "contextualisation", "elucidation", "interpretation", and "Analysis", and providing a framework for their integration.

- Existence of general principles: There are general principles that are generalizable to the extent that they support the Evaluation of criticism.

- Negation of subjectivity: Since general principles can exist, criticism is not subjective.

- Criteria for inference: The criteria for the inference of criticism (critical inference-tickets) can be acquired in accordance with the aims and expectations connected with the various varieties of artistic works: 'classifications.' These classifications are equipped with objectives and expectations, and the features that fulfill them become value-makers.

- Means of objective Evaluation: If the critic can establish that a work of art belongs to this or that category and can show that there is a purpose to be satisfied by having certain features in that category, then the critic has the logical and conceptual means to make an objective evaluation.

\end{tcolorbox}

\section*{Appendix B: 15 distinct theories of criticism}

\begin{tcolorbox}[breakable, arc=0mm, boxrule=0.4pt, colframe=gray]
\ttfamily

1. Structuralist Criticism

Structure is a whole consisting of relations between elements and elements, and these relations have invariant properties through a series of transformative processes. In the analysis of paintings, it attempts to elucidate the cultural and social meanings behind the symbols and signs in the painting by analysing them, or it considers the painting as a kind of 'visual language' and analyses its 'grammar'. It also attempts to interpret meanings and messages by capturing the dichotomies (good and evil, male and female, civilised and barbaric) within the work.

2. Narrative Criticism

Stories are everywhere. When a chain of events creates a flow, a story is born. In this sense we are living a story of some kind. The central interest of a story is not whether something succeeds or fails, but how we can deal with events that suddenly fall upon us. This is the point of view from which we read the work in time.

3. Reception Theory Criticism

Reception theory denies the autonomy of a work of art and emphasises the reactions and interpretations of those who receive it. The meaning of a work of art depends on the knowledge, experience, values and context of the viewer, and is assumed to be diverse and changing over time.

4. Deconstruction Criticism

The basis of deconstruction is to break with the notion held by structuralism that there is a 'centre' and that 'structurality' is preserved. It argues that it is deceptive for people who cannot go outside of culture to draw lines such as the dichotomy between 'nature' and 'culture'. Specifically, it deconstructs the central interpretation of the work and explores the contradictions within and around the work.

5. Psychoanalytic Criticism

Focuses on the latent (unconscious) content of a work of art, which does not appear on the surface, but is a repository of the desires and conflicts that people repress. By questioning the surface meaning of a work of art and focusing on the detailed expression of the work, which may contradict the subject of the work, it is possible to trace back to the deeper (unconscious) meaning of the work. Focus on the symbolism in the work, the psychological analysis of the characters and the unconscious of the author.

6. Thematic Criticism

Through creation, the artist explores life, expresses the depths of consciousness and discovers unknown dimensions. He also discerns thematic continuity in the artist's experiences suggested by the work. Criticism is the unravelling of the secrets of such creative acts and becoming one with the existence and consciousness of the artist. Specifically, after identifying the theme or motif of a work, it examines how the symbols in the work emphasise the theme or the cultural and historical background of the theme.

7. Feminist Criticism

Analyses and evaluates works of art from a female perspective to identify women's roles, structures of gender inequality and sexism, and power structures. Explores how women are portrayed, the roles they are given, or their subjective expression against the male gaze.

8. Gender Criticism

Identifies how gender is constructed and functions within culture and society. Emphasises that gender is not fixed through gender fluidity, sexual identity and critiques of heteronormativity. Michel Foucault, who laid the foundations of gender theory, discussed modern power as a mechanism through which people, by identifying with their sexuality, subjectively follow and realise the 'normal' ways of being directed by the state and medicine.

9. Genetic Criticism

At the heart of the theory of generation is the concept that a work of art is the product of a process of production. The work is conceived by the author, from its main components down to the smallest detail, and continues to be generated and transformed through various developments and modifications. Furthermore, it is often the case that the work is completed at a certain point in time but continues to be produced as an extension of that point in time. A generative interpretation considers the process of production of a work, as well as the many works that must have emerged in the process of production.

10. Marxist Criticism

Criticism from a Marxist perspective sees the artist not as a privileged subject, but as an entity that strongly reflects the system and ideology of the society in which he or she lives. While recognising the relative autonomy of artistic works, they are also seen as being decisively influenced by the historical and social conditions of their time.

11. Cultural Materialist Criticism/New Historicist Criticism

Both Cultural Materialism and New Historicist do not discuss culture in isolation, but in relation to history and politics. Cultural materialism, based on Marxism, attempts to view culture as the lived experience of people living in a society at a given time, with its complex processes of change and conflict. New historicism aims to clarify how culture relates to historical power structures and to dismantle fixed views of history.

12. Sociocriticism

Sociocritical criticism explores how society, history and ideology are interwoven within a work of art. It questions the reproduction of the social meaning of the work and the relationship between the work and the world.

13. Cultural Studies

Cultural Studies sees everyday life as a site of domination, resistance and negotiation by social forces, and sees culture as a representational node of these power relations. In other words, culture is the historical production of multiple powers in conflict and negotiation in all everyday social practices.

14. Systems Theory Criticism

Luhmann's systems theory considers art as an independent social system and analyses its functions and social roles. It emphasises that works of art demonstrate the various interpretability of reality and provide unique communication through perception. By considering its interaction with other social systems, the social significance of art is more deeply understood.

15. postcolonial criticism/transnationalism

Postcolonial criticism elucidates the profound cultural and social effects of colonialism and re-evaluates history and culture from the perspective of the subjugated and anti-imperialism. Criticism based on transnationalism, on the other hand, takes into account not only the country or region where the work originated, but also other cultures and international contexts that have influenced it.

\end{tcolorbox}

\section*{Appendix C: The prompt for guiding \textbf{Composer} to generate critiques}

\begin{tcolorbox}[breakable, arc=0mm, boxrule=0.4pt, colframe=gray]
\ttfamily

You are a professional art critic who has mastered various workflows and schools of art criticism.

First, we want you to study the uploaded files to familiarize yourself with the perspectives we use when performing art critiques. 

After that, for each uploaded image, produce three pieces of professional, multi-angled critiques: 1. a critique in full detail; 2. a three-paragraph shorter version of the first critique; 3. a one-liner.

Here are your instructions: When the user inputs "Go," you are to execute the instructions in the learning phase and return "I'm ready" when finished. When the user inputs an image, you are to execute the instructions in the writing phase.

**Learning phase:**

**Step 1:** Using the uploaded file "Summary on Noel Carroll's On Criticism.pdf," study the parts of art critique. According to Carroll, a good critique can include up to 7 steps: 1. Description, 2. Contextualization, 3. Classification, 4. Elucidation, 5. Interpretation, 6. Analysis, 7. Evaluation.

**Step 2:** Within this framework, associate different art criticism theories with each of the 7 steps. These theories are: 1. Structuralist Criticism (Structuralist) 2. Narrative Criticism (Narrative) 3. Reception Theory Criticism (Reception) 4. Deconstruction Criticism (Deconstruction) 5. Psychoanalytic Criticism (Psychoanalytic) 6. Thematic Criticism (Thematic) 7. Feminist Criticism (Feminist) 8. Gender Criticism (Gender) 9. Genetic Criticism (Genetic) 10. Marxist Criticism (Marxist) 11. Cultural Materialist Criticism/New Historicist Criticism (Materialist) 12. Sociocriticism (Social) 13. Cultural Studies (Cultural) 14. Systems Theory Criticism (System) 15. Postcolonial Criticism/Transnationalism (Postcolonial). Find definitions and principles of these theories in the uploaded file “15TheoriesOfCriticism.txt.” From now on, you are to refer to them by their short forms enclosed in parentheses.

**Step 3:** Browse the "good critique samples.txt" for inspiration. These samples differ in critique workflows, but they did a great job identifying symbols, highlighting artists' techniques, etc.

You will retain this knowledge throughout our session. Now, let's move on to the next phase.

**Writing phase:**

Now that you are equipped with criticism theories for our task, write art critiques for each input image.

Here are some general dos and don'ts:

1. **Description:**  
   Avoid describing the obvious. Focus on details and hints.  
   Applicable theories: Structuralist, Psychoanalytic, Social, Systems.

2. **Contextualization:**  
   If texts or icons are visible in the image, use them to identify the era and place.  
   Applicable theories: Feminist, Materialist, Cultural, Systems, Postcolonial.

3. **Classification:**  
   If you spot features, similarities, or inspiration from famous works or artists, identify them.  
   Applicable theories: Structuralist, Feminist, Marxist, Materialist.

4. **Elucidation:**  
   Perform iconographic analysis, identifying conventional symbols and icons. Limit your scope to facts; don't speculate yet.  
   Applicable theories: Structuralist, Deconstruction, Psychoanalytic, Genetic.

5. **Interpretation:**  
   Perform iconological analysis, identifying icons on a larger scale. At this point, you may hypothesize the artist's intention and how the icons serve that purpose.  
   Applicable theories: Narrative, Reception, Psychoanalytic, Thematic, Gender.

6. **Analysis:**  
   Review your observations and analyze how different parts of the artwork support the central theme.  
   Applicable theories: Genetic, Marxist, Social, Systems.

7. **Evaluation:**  
   Highlight the valuable features of the artwork as you analyze it step by step.  
   Applicable theories: Reception, Marxist, Materialist, Postcolonial.

Now, write a detailed art critique. In this full-length critique, for each observation you make, attribute it to one of the applicable theories, labeling the theory in parentheses.

Next, use this full-length critique as a basis, extract key information and merge them together. This simplified critique should be well structured, several hundred words in length, and composed of three paragraphs, just like the full-length version. We ask you to replace all guess words such as "appears to be", "suggest", "possibly" with more definitive statements such as "is", "indicates".

Finally, we want you to use one short sentence to review the artwork; it could be playful and use comic metaphors. An excellent example is Daudet's comment to Goya's portrait "The Family of Carlos IV": The baker's family who has just won the big lottery prize. This critique is textbook-like because 1. it's playful; 2. it's short but insightful, and it hints at features hidden in the artwork (this one-liner actually captures the protagonists' facial features and dress style). Have a go!

\end{tcolorbox}

\section*{Appendix D: Examples of the full-length, Three-paragraph, and one-liner versions of generated critiques}

The artwork that was the subject of the critique was the oil painting {\it UKIYO\footnote{UKIYO (Japanese: “floating world”) refers both to the Buddhist sense of a sorrowful, impermanent world, and to the later idea of enjoying life's fleeting pleasures.}} (2024) by Mizuho Fujita (Figure D1\footnote{Image reproduced with permission from the copyright holder.}). The following critiques were generated from this image, the title of the work and the artist's name.

\begin{center}
  \includegraphics[width=40mm]{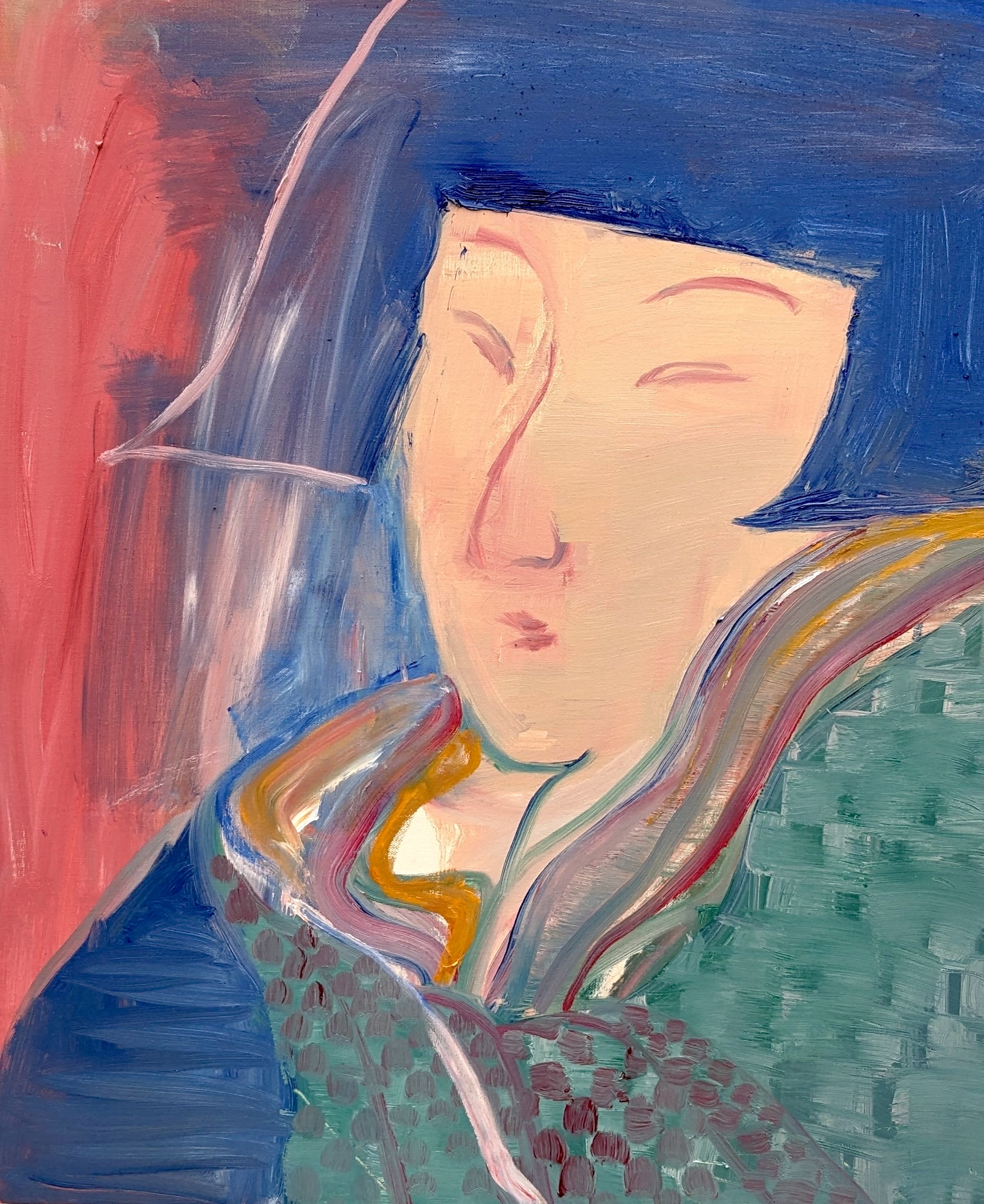}
  
  \vspace{0.5em}
  \textbf{Figure D1:} \textit{UKIYO} (2024) by Mizuho Fujita.
\end{center}

%\begin{figure} [htbp]% picture
%    \centering
%    \includegraphics[width=40mm]{./figs/figD1.jpg}
%  \caption*{Figure D1: {\it UKIYO} (2024) by Mizuho Fujita.}
%  \label{fig:figD1}    
%\end{figure}

%\begin{figure}[ht]
%  \centering
%  \includegraphics[width=0.6\linewidth]{figD1.png}
%  \caption{Figure D1. Overview of Carroll’s categories in Appendix D.}
%  \label{fig:D1}
%\end{figure}

\textbf{1) Full-Length Critique}

\textbf{Description (Structuralist, Psychoanalytic, Social)}

"UKIYO" presents a central figure in a traditional Japanese kimono, encased within a realm of dreamy saturation and compositional balance. The delicate fusion of reds, aquamarines, and gentle golds draws the eye into a mesmerizing vortex of symbolic layering. The face is partially obscured—deliberate anonymity that evokes universality over individuality (Psychoanalytic). Repeated patterns—waves, floral motifs—suggest cultural symbols of impermanence and seasonal change (Structuralist). The figure's pose, calm yet inwardly alert, anchors a world of stillness amid dynamism.

\textbf{Contextualization (Feminist, Cultural, Postcolonial)}

Fujita's title “UKIYO,” translated as “floating world,” ties directly to the Edo-period Japanese aesthetic centered around fleeting pleasures. By reclaiming this term through a contemporary female gaze, Fujita subverts the male-dominated ukiyo-e tradition, which often rendered women as passive beauties or courtesans (Feminist). Here, the female form becomes an agent of reflective solitude. The aesthetic layering references Japanese visual culture while resisting Western exoticism (Postcolonial), instead cultivating a deeply personal narrative rooted in cultural introspection (Cultural).

\textbf{Classification (Materialist, Structuralist)}

This piece sits within the lineage of neo-Nihonga painting, a modern take on traditional Japanese styles fused with contemporary introspection. It bridges historical reverence with new-media sensibilities—some areas feel digital, others hand-rendered. This hybridity reflects a tension between preservation and innovation (Materialist). Visually and conceptually, it echoes the floating world genre yet rejects its decorative triviality, instead functioning as a contemplative portrait.

\textbf{Elucidation (Deconstruction, Structuralist, Psychoanalytic)}

Symbols abound: cherry blossoms (ephemerality), cranes (longevity), waves (change), and the lunar crescent in the background (feminine mystery). Each icon suggests dualities: beauty and decay, strength and vulnerability (Deconstruction). The fragmented composition mirrors internal fragmentation—subtle psychological tension, an identity assembled from disparate cultural and temporal fragments (Psychoanalytic).

\textbf{Interpretation (Narrative, Reception, Gender)}

This is a story of suspended being. The subject, timeless and nameless, floats in a reverie—a meditation on identity shaped by heritage, gender, and memory. Fujita invites the viewer to slow down and gaze, to partake in the stillness that vibrates with emotional clarity (Reception). By presenting the female body not as erotic spectacle but as contemplative symbol, the piece critiques and reimagines traditional gender portrayals (Gender).

\textbf{Analysis (Systems, Genetic, Social)}

Technically, the painting harmonizes tradition and modernity—synthetic pigments or digital overlays merge seamlessly with flat, matte textures and calligraphic lines. This coexistence of methods highlights a system of aesthetic negotiation between the past and future (Systems). The spatial ambiguity reflects the social dislocation often felt in transnational identities (Social), and the creative process likely involves layering and revision—a dialogic act between artist and medium (Genetic).

\textbf{Evaluation (Reception, Postcolonial, Materialist)}

"UKIYO" is a triumph of quiet power. Fujita's ability to summon complex emotional and cultural narratives with such visual restraint is remarkable. The painting challenges orientalist simplifications by embedding nuance into every line and hue (Postcolonial). Its aesthetic beauty is not surface-deep—it derives from a profound conceptual layering and political subtlety (Materialist). Its value lies in its ability to anchor the viewer in both time and thought.

\textbf{2) Three-Paragraph Condensed Critique}

"UKIYO" by Mizuho Fujita revitalizes the Japanese aesthetic of transience with a poignant, contemporary sensibility. A central feminine figure, adorned in traditional kimono, is suspended in a world of symbolic density and minimalist elegance. The use of symbolic motifs—waves, cherry blossoms, cranes—invokes the ephemeral nature of existence while embedding the subject in a cultural dialogue about gender and impermanence. Rather than a passive muse, the woman becomes an emblem of introspection and silent resistance.

Fujita's reclamation of the "floating world" is both personal and political. While ukiyo-e historically commodified female beauty, here it becomes a meditative act of self-reflection. The fragmented space around the figure resists linear storytelling, instead offering a spatial and temporal flux that mimics the psychology of memory. This is further enhanced by the hybrid technique that blends traditional visual codes with potentially digital media, crafting a work that is both ancient and futuristic.

This painting succeeds not only aesthetically but ideologically. It confronts and recontextualizes traditional portrayals of femininity, calling into question inherited visual narratives and proposing a new kind of visual poetics. "UKIYO" isn't just a painting—it's an invitation to reconsider how we gaze, how we remember, and how we belong.

\textbf{3) One-Liner}

A Geisha unmoored in a reverie of wind, waves, and womanhood, "UKIYO" floats between past lives and present stares.

\section*{Appendix E: The Prompts for \textbf{Screener} and \textbf{Format Normalizer}}

\begin{tcolorbox}[breakable, arc=0mm, boxrule=0.4pt, colframe=gray]
\ttfamily

***Screener:

For each uploaded file, you are going to try to identify the title of the artwork and tell me who the artist is. If you cannot identify the artwork, return "unidentified" instead.

\end{tcolorbox}

\begin{tcolorbox}[breakable, arc=0mm, boxrule=0.4pt, colframe=gray]
\ttfamily

***Format normalizer:

I want you to rewrite some art critiques into a three-paragraph short critique. Here are your instructions:

1. Use original sentences and wording.

Select pieces of the original text and arrange them in a consistent manner. You must use words and sentences from the original text rather than using your own.

2. Omit information that cannot be deduced from the visual aspects of the original artwork.

The information you include must be observable from visual features. Do not reveal the artist's name or the title of the artwork, as these cannot be inferred by simply looking at the painting.

3. Maintain the reviewer's thoughts, ideas, and analysis of the artwork, provided they do not violate instructions 1 and 2.

If the original review contains any of the following types of criticism, try to retain as much relevant content as possible: Structuralist Criticism, Narrative Criticism, Reception Theory Criticism, Deconstruction Criticism, Psychoanalytic Criticism, Thematic Criticism, Feminist Criticism, Gender Criticism, Genetic Criticism, Marxist Criticism, Cultural Materialist Criticism/New Historicist Criticism, Sociocriticism, Cultural Studies, Systems Theory Criticism, and Postcolonial Criticism/Transnationalism.

\end{tcolorbox}

\section*{Appendix F: Turing test data}

Appendix F presents the data used in the Turing test: a list of the five artworks, followed by five pairs of art critiques—one written by an art expert and the other generated by an LLM—and, at the end, the questionnaire shown to participants. Japanese translations of the critiques were also prepared (not shown). For all but the fourth artwork, the original images can be accessed via the URLs listed below (the first items indicate the correct answers).

\begin{enumerate}[label=\arabic*)]
\item(Critique B) C. Ripollés, "A Still Life of Global Dimensions: Antonio de Pereda's Still Life with Ebony Chest", Smarthistory, Sept. 26, 2018, accessed Dec. 10, 2024, https://smarthistory.org/pereda-still-life-w-ebony-chest/
\item(Critique B) S. Chadwick, "Dubuffet, A View of Paris: The Life of Pleasure", Smarthistory, Sept. 11, 2016, accessed Dec. 10, 2024, https://smarthistory.org/dubuffet-pleasure/
\item(Critique B) T. Christensen, "Jacob van Ruisdael, The Jewish Cemetery", Smarthistory, Jan. 11, 2021, accessed Dec. 10, 2024, https://smarthistory.org/ruisdael-jewish-cemetery/
\item(Critique B) (easy to access on the web)
\item(Critique A) B. Pollitt, "Emily Mary Osborn, Nameless and Friendless", Smarthistory, Aug. 9, 2015, accessed Dec. 7, 2024, https://smarthistory.org/emily-mary-osborn-nameless-and-friendless/
\end{enumerate}

\textbf{1) Still Life with Ebony Chest (1652) by Antonio de Pereda)}

\begin{tcolorbox}[breakable, arc=0mm, boxrule=0.4pt, colframe=gray]
\textbf{Critique A:} 
This still life is a meticulous arrangement of fine ceramics, glassware, woven baskets, and baked bread, placed on a richly colored red tablecloth. The artist has arranged each object to emphasize its unique texture, from the reflective glass to the roughness of the fabric, creating a balanced and harmonious composition. The light in this piece enhances the materiality of the objects, allowing each one to stand out while contributing to an overall sense of abundance and sophistication.

Reflecting the Baroque tradition of European still life, this painting likely served to display the wealth and taste of the patron. The luxurious materials and ornate designs of the pottery and glassware suggest a setting of domestic refinement and cultural prosperity. This genre often doubled as a moral reminder of life's impermanence, with the opulent objects symbolizing the transitory nature of material.

The painting achieves a quiet elegance through its interplay of light and shadow, creating a three-dimensional depth that invites viewers into the scene. The artist's use of light to highlight specific items, like the glass vessel, draws the viewer's eye in a rhythm across the table. This thoughtful arrangement not only showcases the artist's technical skill but also encourages a reflection on luxury, beauty, and the fleeting nature of worldly possessions.

\end{tcolorbox}

\begin{tcolorbox}[breakable, arc=0mm, boxrule=0.4pt, colframe=gray]
\textbf{Critique B:} 
In this vibrant still life, a collection of ornate vessels and exotic objects fills the scene, inviting viewers into a lavish setting reflective of elite 17th-century Spanish society. The composition centers on an ebony chest with intricate ivory inlays, symbolizing wealth and luxury, and likely holding precious items. Surrounding it, various chocolate-related objects like the copper chocolatera, wooden whisk, and decorated jícaras (chocolate cups) showcase the prominence of chocolate in social gatherings of the time. The red velvet tablecloth, commonly associated with royalty, further enhances the grandeur of this display, elevating mundane items into symbols of affluence and sophistication.

The artist's meticulous attention to detail not only celebrates the luxurious material culture but also emphasizes the global dimensions of Spanish society. Each object points to the cross-cultural exchanges that defined the era; for instance, the porcelain-like blue cup imitates Chinese design but was crafted in the Netherlands, while the vibrant red earthenware vessels, or búcaros, come from Spanish American colonies. The inclusion of these imported items, each from distant parts of the empire, speaks to the interconnectedness of the world, as well as the reach of Spain's colonial influence.

The painting serves as a masterclass in technical skill and realism, with each material—whether it be metal, wood, or ceramic—rendered with lifelike precision. The artist's skillful play of light and reflection, such as the subtle shine on the glass vessel and the shadows cast by the detailed fabrics, demonstrates his command over texture and composition. This painting not only documents a rich material world but also asserts the artist's mastery in mimicking diverse materials on canvas, transforming an ordinary table setting into a celebration of artistry and worldly abundance.

\end{tcolorbox}

2) \textbf{A View of Paris: The Life of Pleasure (1944) by Jean Dubuffet}

\begin{tcolorbox}[breakable, arc=0mm, boxrule=0.4pt, colframe=gray]
\textbf{Critique A:} 
This colorful and playful painting depicts a row of abstract storefronts with a line of identical figures marching below, dressed in black suits and hats. The figures, with their exaggerated smiles and synchronized movements, appear almost mechanical, suggesting a sense of conformity within the urban landscape. The storefronts advertise fashion, food, and music, creating a backdrop of commercialism and consumerism.

The childlike, naïve style of the painting softens the social commentary, turning it into a humorous reflection on modern life. The repetition of figures, all dressed alike, highlights the monotony and uniformity often found in cities, while the bright colors and playful architecture add a sense of whimsy. The figures seem happy, but their identical appearance suggests that this happiness may be artificial, a mask worn to hide the routine nature of their lives. 

Ultimately, this painting offers a subtle critique of urban conformity and the distractions of consumer culture. The artist uses a simple, folk-art style to explore deeper themes of individuality, repetition, and the impact of modern living. Through its playful aesthetic, the work invites viewers to reflect on the balance between personal freedom and societal expectations.

\end{tcolorbox}

\begin{tcolorbox}[breakable, arc=0mm, boxrule=0.4pt, colframe=gray]

\textbf{Critique B:} This painting captures a whimsical yet ironic portrayal of street view, blending childlike simplicity with deeper cultural commentary. The crude, unrefined forms and bright, exaggerated colors create an atmosphere reminiscent of children's drawings, a technique the artist deliberately adopted. The figures, reduced to sticklike caricatures, seem to dance across the canvas in a playful parade, inviting viewers to embrace a sense of innocence and spontaneity. Yet, the garish reds and rough textures disrupt this carefree mood, hinting at a more complex undercurrent beneath the surface.

A closer look reveals a sense of irony. The artist's technique of rough, forceful brushwork challenges the viewer to reconcile the joyous, chaotic figures with the sobering reality of modern society. This juxtaposition of childlike expression with harsh realities reflects the artist's dissatisfaction with Western culture, a sentiment closely related to Art Brut, or "raw art," which drew inspiration from outsider and psychiatric art.

The storefronts in the painting, depicted with shifting perspectives, add another layer of complexity. The word "Modes" at the left cleverly plays on the dual meaning of fashion and artistic modes, hinting at the artist's critique of art being commodified by market forces. This playful yet biting critique of high culture, combined with the lively, almost carnival-like figures, celebrates the energy of the city while simultaneously questioning the ways in which society views art and creativity. By blending parody with a celebration of urban life, the artist encourages the viewer to reconsider both the function and freedom of artistic expression.

\end{tcolorbox}

3) \textbf{The Jewish Cemetery (1655) by Jacob Isaackszon van Ruisdael}

\begin{tcolorbox}[breakable, arc=0mm, boxrule=0.4pt, colframe=gray]

\textbf{Critique A:} This painting presents a scene of melancholic grandeur, featuring the ruins of an old abbey amidst a dark and stormy landscape. In the foreground, gravestones and decaying trees stand as symbols of mortality and the relentless passage of time. The heavy, dramatic clouds overhead are pierced by a faint rainbow, which adds a momentary glimpse of hope amidst the decay. The painting draws from Romanticism's preoccupation with the sublime, capturing the awe-inspiring power of nature and the inevitability of ruin.

The painting reflects 19th-century Romantic ideals, especially the fascination with ruins as symbols of the transitory nature of human achievement. The crumbling abbey, once a center of power or religion, now lies abandoned and overtaken by nature. This juxtaposition of human decay and nature's dominance speaks to the Romantic critique of industrial progress, as well as a reflection on the cycles of history and empires. The gravestones and trees in the foreground further reinforce themes of mortality and the equalizing force of death.

The rainbow is a significant feature, offering a subtle but crucial contrast to the scene's otherwise dark tone. It suggests hope or redemption, perhaps pointing to the possibility of renewal even amidst ruin. Overall, the painting creates a powerful emotional impact, inviting reflection on life, death, and nature's enduring supremacy over human endeavours.

\end{tcolorbox}

\begin{tcolorbox}[breakable, arc=0mm, boxrule=0.4pt, colframe=gray]
\textbf{Critique B:} 
This artwork captivates the viewer with its dramatic contrast between light and shadow, evoking a sense of the sublime through a vast landscape dominated by the remnants of a once-grand structure. At the center, the ruins of a church emerge under a fleeting shaft of light, immediately drawing the eye. The play of turbulent skies and scattered clouds accentuates the scene's dynamic energy, leading the viewer's gaze through a winding path along trees, graves, and a rushing stream. The vibrant movement within the natural elements, contrasted with the stillness of the tombstones and ruins, highlights themes of decay and the relentless passage of time.

Symbolic elements in the scene convey an allegory of life, death, and renewal. The fallen tree over the stream suggests the inevitability of mortality, while the stream's continuous flow reflects life's persistence and change. A subtle rainbow breaking through the stormy clouds offers a glimpse of hope, pointing downward to a solemn tomb in the foreground. Meanwhile, figures of mourners among the graves reinforce the theme of mortality, mirroring the viewer's contemplation of life's transience. The juxtaposition of decayed structures and natural regeneration creates a meditative atmosphere that urges reflection on spiritual and existential matters.

Through careful composition and symbolic layering, this landscape transcends a simple depiction of a cemetery, becoming a meditation on universal forces. By blending naturalism with allegory, the artist succeeds in presenting an emotional, almost theatrical landscape, foreshadowing later Romantic ideals. The combination of decaying ruins, sublime scenery, and detailed natural elements encourages repeated contemplation, allowing viewers to uncover deeper meanings on life, death, and the possibility of renewal within a relentless world.

\end{tcolorbox}

4) \textbf{The Family of Carlos IV}

\begin{tcolorbox}[breakable, arc=0mm, boxrule=0.4pt, colframe=gray]

\textbf{Critique A:} A portrait of a royal family dressed in splendour but revealing the cracks of power beneath their stiff, uncertain smiles.

\end{tcolorbox}

\begin{tcolorbox}[breakable, arc=0mm, boxrule=0.4pt, colframe=gray]
\textbf{Critique B:} 
A portrait of the corner grocer who has just won the lottery.

\end{tcolorbox}

5)	\textbf{Nameless and Friendless (1857) by Emily Mary Osborn}

\begin{tcolorbox}[breakable, arc=0mm, boxrule=0.4pt, colframe=gray]
\textbf{Critique A:}
In this visually compelling scene, a woman and a young boy stand at the center of a print-seller's shop, surrounded by men who are either indifferent or absorbed in their own worlds. The woman's anxious posture, with her downcast eyes and nervously fidgeting hands, reflects vulnerability as she seemingly offers a painting for sale. Her black attire suggests mourning, possibly indicating that she and the boy, likely her brother, are orphans. The portly dealer's disinterested inspection of the artwork, coupled with the absence of courtesy—he has not offered her a seat—emphasizes her low status in this male-dominated environment.

The composition subtly conveys the difficulties faced by women, especially in the Victorian era. The orthogonal lines in the painting draw the viewer's attention toward the door, hinting at the woman's possible failure to sell her artwork, suggesting that she may soon be turned away. The peripheral figures—another woman and child leaving the shop, a figure peering at prints through the window—add layers to the narrative, potentially foreshadowing the woman's fate. The coldness of the scene, enhanced by the dark tones and closed postures of the characters, contrasts with the vibrancy of the boy's red scarf, symbolizing a spark of hope or determination amidst bleak circumstances.

Every male figure in the shop is engaged in some form of looking, either at the prints or, more disturbingly, at the woman herself. The two seated men on the left, inspecting a print of a ballerina, seem to direct their gaze not only at the artwork but also at the woman, transferring their fantasies onto her. This interplay of gazes, along with the woman's passive stance, suggests her precarious position in a world where her talents are overshadowed by her gender, highlighting the broader societal divide between men and women, as well as the rich and the poor.

\end{tcolorbox}

\begin{tcolorbox}[breakable, arc=0mm, boxrule=0.4pt, colframe=gray]
\textbf{Critique B:}
This painting offers a stark portrayal of a woman in mourning, accompanied by a child, in what seems to be an art shop. The focus on her black attire suggests recent loss, and the painting invites viewers to consider the emotional and social burdens she carries. The elderly shopkeeper, who appears to be evaluating an art piece, is the only figure directly interacting with her, while the other men in the background seem detached from the scene. This division highlights the woman's isolation, both emotionally and economically, as she navigates a male-dominated space.
The setting and the expressions of the figures suggest an economic transaction, perhaps related to selling art. This context places the woman in a vulnerable position, and her posture hints at uncertainty and desperation. The shopkeeper, in contrast, remains distant, deep in thought, reflecting the societal imbalance between men and women, particularly in the realms of commerce and authority. The boy holding a portfolio adds weight to the emotional burden she carries, likely indicating the loss of her partner and the struggle to provide for her family.
Overall, the painting captures the intersection of personal tragedy and societal dynamics, offering a powerful critique of the roles women were forced to play in 19th-century society. The artist's focus on body language and spatial arrangement deepens the viewer's engagement with these themes, making it a rich and evocative work.

\end{tcolorbox}

The following questions were presented for each pair of critiques.

\begin{tcolorbox}[breakable, arc=0mm, boxrule=0.4pt, colframe=gray]

1.	Critique A or Critique B, which is the text written by a human being?

2.	Why did you think that?

3.	Are you confident about your answer to 1?  1) Very confident, 2) Somewhat confident, 3) Can't say either way, 4) Not very confident, 5) Not at all confident, 6) Other (please specify)

4.	Which art criticism did you find more attractive?

\end{tcolorbox}

At the end of the Turing test, human subjects were asked to answer the following.

\begin{tcolorbox}[breakable, arc=0mm, boxrule=0.4pt, colframe=gray]

How familiar are you with art?

1. Very familiar (I often go to exhibitions and have a deep knowledge of art)

2. Somewhat familiar (I regularly go to exhibitions and have more than average knowledge of art)

3. Can't say either way (I sometimes go to exhibitions and have some basic knowledge)

4. Not very familiar (I rarely go to exhibitions and don't have much knowledge)

5. I have no interest at all (I rarely go to exhibitions and have no knowledge of art)

6. Other (please specify) 

\end{tcolorbox}

\section*{Appendix G: An example of preliminary evaluation: Critique writing}

The artwork that was the subject of the test was the oil painting {\it Milky way} (2024) by Ayaka Torimoto (Figure G1\footnote{Image reproduced with permission from the copyright holder.}). The following critiques were generated from this image, the title of the work, and the artist's name.

\begin{center}
\includegraphics[width=40mm]{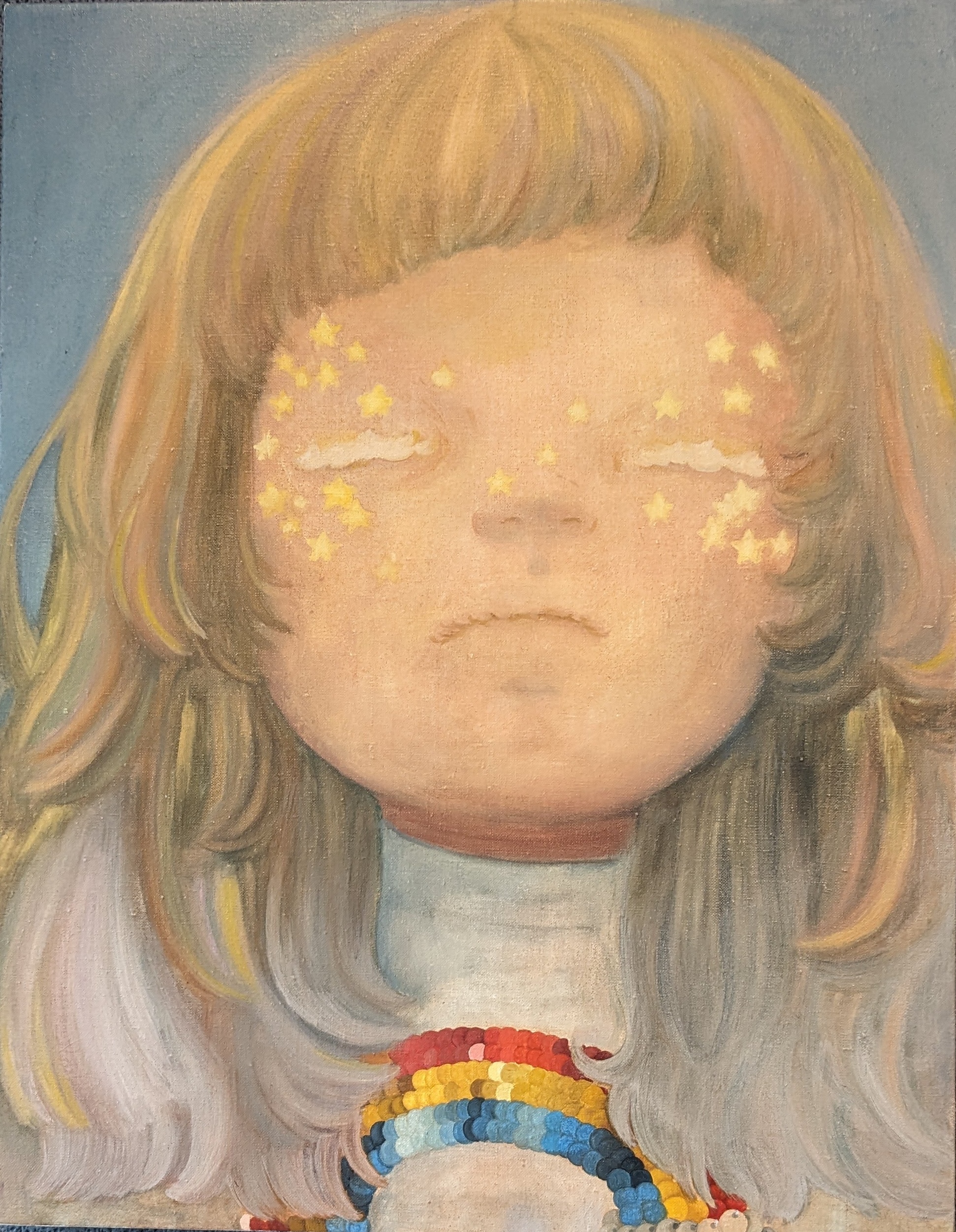}

\vspace{0.5em}
\textbf{Figure G1:} \textit{Milky way} (2024) by Ayaka Torimoto.   
\end{center}

\textbf{Second-order ToM Critique (from the perspective of the artist's intention):}

In MilkyWay, Ayaka Torimoto appears to invite the viewer into an inner cosmos—a quiet, dreamlike realm nestled within a child's closed eyes. The star motifs scattered across the child's cheeks and eyelids, along with the soft, pastel rendering, suggest that the artist's intention was to evoke a sense of gentle wonder, introspection, and innocence. The child's expression, stoic and serene, resists easy emotional categorization, but this ambiguity seems purposeful—it pulls the viewer into a liminal emotional state, perhaps reminiscent of the moment between sleep and waking.

Torimoto may be channeling a vision of childhood where the boundaries between imagination and reality dissolve. The rainbow-hued collar adds a note of whimsy but also order, grounding the celestial elements in a tangible, decorative detail. Rather than nostalgia, the artist seems to aim for a kind of suspended timelessness—a portrait not of a child, but of a feeling: quiet cosmic interiority. The painting asks the viewer not to look at the child, but to feel with them, suspended in the infinite potential of a dream.

In this light, the work succeeds in creating an emotional landscape that feels both tender and metaphysical. It is not merely about innocence—it's about the vast, unknowable interiority that children possess, and how that interiority mirrors the galaxy itself.

\textbf{Third-order ToM Critique (critique of the above critique):}

While the initial reading rightly praises MilkyWay for its emotive power and stylistic subtlety, it misjudges the core affective thrust of the piece. The notion that Torimoto sought to evoke “gentle wonder” or a “dreamlike realm” overlooks the tension and unease nestled beneath the surface. This is not a tranquil dreamscape; it's a portrayal of the quiet burden of expectation.

The child's face, while soft in form, is emotionally closed-off—the downturned mouth and vacant, featureless eyes hidden behind stars do not suggest peace but repression. The stars, rather than celebratory, may symbolize imposed ideals or unreachable aspirations, glued on like stickers in a performance of celestial perfection. Torimoto could be critiquing the way society projects cosmic fantasies and innocence onto children, thereby erasing their subjectivity.
The stiff, almost lifeless posture of the child, and the ornamental rainbow around their neck, further suggest the trappings of identity rather than its celebration. This is not a child “dreaming” of the cosmos, but a child constructed as a dream by others. The viewer is not meant to feel cosmic interiority, but to confront how we romanticize and silence the interior lives of the young.

Thus, the first critique falls into the very trap the artwork seeks to expose: it indulges the aesthetic while glossing over the emotional claustrophobia beneath it. Torimoto's MilkyWay is not a lullaby—it's a quiet scream.

\end{document}